\documentclass[11pt,a4paper,reqno]{amsart}

\usepackage[margin=1in]{geometry}  
\usepackage[utf8]{inputenc}
\usepackage{amsmath}
\usepackage{amsfonts}
\usepackage{amsthm}
\usepackage{mathtools}
\usepackage{yhmath}
\usepackage{hyperref}
\usepackage{graphicx}

\usepackage{libertine}
\usepackage{libertinust1math}
\usepackage[T1]{fontenc}

\usepackage{caption}
\DeclareCaptionFont{libertine}{\footnotesize}
\usepackage{booktabs} % better tabular
\usepackage{tabularx}
\usepackage{colortbl}
\usepackage{multirow}
\usepackage{hyphenat}
\usepackage{mathrsfs}
\usepackage{mathtools}
\mathtoolsset{centercolon}
\usepackage{nicefrac}
\usepackage{units}

\usepackage[backend=biber, style=numeric-comp, natbib=true, maxbibnames=6, firstinits=true]{biblatex}

\bibliography{References.bib}
\appto{\bibsetup}{\sloppy} %assure line-break in references
\usepackage{cleveref}
\crefmultiformat{subequation}%
{\edef\crefstripprefixinfo{#1}(#2#1#3}%
{,#2\crefstripprefix{\crefstripprefixinfo}{#1}#3)}%
{,#2\crefstripprefix{\crefstripprefixinfo}{#1}#3}%
{,#2\crefstripprefix{\crefstripprefixinfo}{#1}#3)}

\newcommand{\figref}[1]{\textup{Fig.~\ref{#1}}}

\newcommand{\teqref}[1]{\textup{Eq.~(\ref{#1})}}

\newcommand{\secref}[1]{\textup{Section~\ref{#1}}}

\newcommand{\suppsecref}[1]{\textup{Section~\ref{#1}}}

\newcommand{\figloc}[1]{\emph{#1}}
\def\cf{\emph{cf.}}
\def\ie{\emph{i.e.}}
\def\eg{\emph{e.g.}}

\def\resp{resp.}

\def\id{\operatorname{id}}

\DeclareMathOperator*{\argmin}{argmin}

\DeclareMathOperator{\SO}{SO}

\DeclareMathOperator{\SE}{SE}

\newcommand{\Hess}{\mathop{\mathrm{Hess}}\nolimits}

\def\ext{\operatorname{ext}}

\def\gam{{\gamma}}
\def\hgam{{\hat{\gam}}}

\def\ext{{\rm out}}
\def\inn{{\rm in}}

\def\dext{{\texttt{out}}}
\def\dint{{\texttt{in}}}
\def\ben{{\texttt{ben}}}
\def\mem{{\texttt{mem}}}

\def\cWben{{\cW_{\rm ben}}}
\def\cWmem{{\cW_{\rm mem}}}
\def\wben{{\sigma_{\rm ben}}}
\def\wmem{{\sigma_{\rm mem}}}

\def\dTH{{g_{\rm T\&H}}}

\def\cA{\mathcal{A}}
\def\cB{\mathcal{B}}
\def\cM{\mathcal{M}}
\def\cE{\mathcal{E}}
\def\cS{\mathcal{S}}
\def\cH{\mathcal{H}}
\def\cW{\mathcal{W}}
\def\RR{\mathbb{R}}
\newcommand{\dt}{{\mathrm{d}t}}
\def\gam{{\gamma}}
\def\hgam{{\hat{\gam}}}

\def\ext{{\rm out}}
\def\inn{{\rm in}}

\def\dext{{\texttt{out}}}
\def\dint{{\texttt{in}}}
\def\ben{{\texttt{ben}}}

\def\cWben{{\cW_{\gam}^{\rm bend}}}
\def\cWmem{{\cW_{\gam}^{\rm strain}}}
\def\wben{{\sigma_{\rm bend}}}
\def\wmem{{\sigma_{\rm strain}}}

\def\dTH{{g_{\rm T\&H}}}
\def\sfM{\mathsf{M}}
\def\sfS{\mathsf{S}}
\def\sfB{\mathsf{B}}
\def\sfE{\mathsf{E}}
\def\sfW{\mathsf{W}}
\def\gamtk{{\gam_{t_k}}}
\def\gamtkk{{\gam_{t_{k+1}}}}
\def\tk{{t_k}}
\def\tkk{{t_{k+1}}}
\def\dext{{\texttt{out}}}
\def\dint{{\texttt{in}}}
\def\ben{{\texttt{bend}}}
\def\mem{{\texttt{strain}}}

\begin{document}
\title{\sc Sub--Riemannian boundary value problems for Optimal Geometric Locomotion}

\author{Oliver Gross} 
\address{University of California, San Diego, 9500 Gilman Dr, La Jolla, CA 92093, USA}
\curraddr{}
\email{ogross@ucsd.edu}

\author{Florine Hartwig} 
\address[A1]{University of Bonn, Endenicher Allee 60, 53115, Bonn, Germany}
\curraddr{}
\email{florine.hartwig@uni-bonn.de}

\author{Martin Rumpf}
\address{University of Bonn, Endenicher Allee 60, 53115, Bonn, Germany}
\curraddr{}
\email{martin.rumpf@uni-bonn.de}
\thanks{}

\author{Peter Schr\"oder}
\address{California Institute of Technology, 1200 E California Blvd, Pasadena, CA 91125, USA\newline\indent University of Bonn, Endenicher Allee 60, 53115, Bonn, Germany}
\curraddr{}
\email{ps@caltech.edu}
\thanks{}

\begin{abstract}
	We propose a geometric model for optimal shape-change-induced motions of slender locomotors, \eg, snakes slithering on sand. In these scenarios, the motion of a body in world coordinates is completely determined by the sequence of shapes it assumes. Specifically, we formulate Lagrangian least-dissipation principles as boundary value problems whose solutions are given by sub-Riemannian geodesics. Notably, our geometric model accounts not only for the energy dissipated by the body's displacement through the environment, but also for the energy dissipated by the animal's metabolism or a robot's actuators to induce shape changes such as bending and stretching, thus capturing overall locomotion efficiency. Our  continuous model, together with a consistent time and space discretization, enables numerical computation of sub-Riemannian geodesics for three different types of boundary conditions, \ie, 
	fixing initial and target body, 
	restricting to cyclic motion, or
	solely prescribing body displacement and orientation. 
	The resulting optimal deformation gaits qualitatively match observed motion trajectories of organisms such as snakes and spermatozoa, as well as known optimality results for low-dimensional systems such as Purcell's swimmers. Moreover, being geometrically less rigid than previous frameworks, our model enables new insights into locomotion mechanisms of, \eg, generalized Purcell's swimmers. The code is publicly available.
\end{abstract}

\maketitle

\noindent Understanding how animals navigate and propel themselves remains a fundamental question in biomechanics~\cite{Dickinson:2000:HAM}.
The ability to maneuver robustly and efficiently has substantially influenced designs in fields such as mechanical engineering and robotics~\cite{Hirose:2009:SLR, Chong:2022:CTL, Chikere:2025:EDE}.
Locomotion dynamics are tightly coupled to environmental conditions. Complex interactions between an organism's morphology and its surrounding medium shape key performance metrics such as speed, maneuverability, and energy efficiency~\cite{Schiebel:2019:MDR, Guasto:2020:FKR, Yaqoob:2023:MOU}. While many legged creatures move by walking and/or crawling, animals with poorly developed or no legs have established a number of entirely different strategies~\cite{Gray:1955:PSU, Lighthill:1969:HAA, Purcell:1977:LLR, Dickinson:2000:HAM, Maladen:2009:USS, Chong:2022:CTL}. 

Notably, a number of organisms achieve locomotion solely by changing their body shape~\cite{Quillin:1999:KSL, Hu:2009:MSL,  Lauder:2015:FLR, Rieser:2024:GPp}. For example, earthworms move through peristaltic body deformations, where waves of muscle contractions propagate along the body to produce forward motion. Snakes, which employ a variety of gaits such as lateral undulation (or \emph{slithering}), concertina movement, sidewinding, and rectilinear locomotion~\cite{Gray:1946:MLS, Jayne:1986:KTSL}, arguably represent the most prominent example.

Motion due to shape changes can be derived from fundamental geometric
principles~\cite{Purcell:1977:LLR, Shapere:1989:GKD, Kanso:2005:LAB,
	Glisman:2022:SPF}. In its simplest form, it is modeled based on
interactions between the body and the environment, dominated by
\emph{outer dissipation}. An example is the viscous friction 
experienced by a body moving through a low Reynolds number fluid~\cite{Lighthill:1969:HAA}.
The slithering locomotion of snakes on sand is based on similar
principles. Due to a frictional anisotropy of their ventral scales,
displacements of their body in the normal direction dissipate
considerably more energy than those in the tangential direction. This
anisotropy enables net propulsion, which otherwise would not be
possible~\cite{Gray:1955:PSU, Lighthill:1969:HAA}: Much like the
forward motion generated by pushing the blades of an ice skate
in the near normal direction. In this way the environment exerts resistive
forces on the body, directed opposite to the normal
components of the shape change velocities and weighted by the
anisotropy. On surfaces with sufficient protrusions, such as grass,
gravel, or sand, this results in snakes propelling themselves with undulations.

Comprehensive theoretical analysis of the underlying dynamical systems necessitates suitable models of the dynamics. 
So-called \emph{geometric mechanics}~\cite{Marsden:1999:IMS} has found particularly successful application in 
the study of locomotive systems in both biomechanics~\cite{Shapere:1989:GSP, Rieser:2024:GPp} and robotics~\cite{Ostrowski:1998:GMU, Hatton:2015:NNL, Yang:2024:TGM}. It  builds on a formal distinction between shapes and their respective positioning in world space 
to express the dynamic conservation laws of Newtonian mechanics in terms of geometric constraints. 

Furthermore,  the abstract framework of \emph{sub-Riemannian
	geometry}~\cite{montgomery:2002:ATo}---a generalization of
Riemannian geometry that considers configuration spaces subject to
linear constraints---addresses questions concerning the
\emph{controllability} of such systems, \ie, the existence of
physically realizable motion trajectories from one configuration, say,
a snake shape positioned in world space, to another---so-called ``parallel parking'' problems~\cite{Frankel:2011:GOP}.
Necessary and sufficient conditions for the existence of such paths
are addressed by the so-called Chow--Rashevskii theorem~\cite[Sec. 1.6]{montgomery:2002:ATo}.
However, rigorously verifying its assumptions is typically nontrivial and requires case by case investigation. 
Previous works have taken on this task and established controllability
for a number of kinematic locomotors~\cite{Alouges:2008:OSL,
	Giraldi:2013:COS, Kadam:2016:CPS, Moreau:2023:COC}, while
theoretical limits on the achievable travel distance have been studied under suitably chosen simplifying assumptions in~\cite{ Kuznetsov:1967:MMS,Becker:2003:SPM,Tam:2007:OSP,Giraldi:2015:ODP}.

Supposing paths between any two configurations exist, the question
arises how to select a particular path. This could be based on
minimizing the outer dissipation integrated along the
trajectory~\cite{Hatton:2012:KKL, Hatton:2017:KCE,
	Becker:2025:IGL}.
Given nature's tendency toward resource
efficiency, locomotion strategies that optimize such notions of efficiency are
commonly regarded as plausible explanations for observed behavioral
patterns. There are a number of approaches in the literature to
capture notions of \emph{effective} locomotion~\cite{Shapere:1989:ESP, Becker:2003:SPM, Tam:2007:OSP,Chambrion:2019:OSD}. In \cite{Lighthill:1969:HAA} and \cite{Purcell:1977:LLR}, authors already characterized efficiency in terms of energy expenditure. 
More contemporary approaches instead compare the total energy expenditure with the power required to drag the locomotor's body in a reference shape the same distance at the same average speed~\cite{Becker:2003:SPM, Tam:2007:OSP}.

If the dynamics of locomotion can be expressed in geometric terms then
so should be notions of optimality.
In this paper we employ sub-Riemannian geometry, \cf~\cite{Ramasamy:2019:GOG}, to achieve this goal.
Additionally we augment the outer dissipation with an \emph{inner
	dissipation} term, built into the geometric formulation of
locomotion. Only recently has the importance of accounting for
additional inner energy expenditures been
recognized~\cite{Hatton:2022:GOG}.
This inner dissipation accounts for the energetic cost of
bending and stretching/compression deformations of the body. 
These include, among others, cases such
as the peristaltic motion of worms, segment-driven telescoping in soft robots,
slithering motions of snakes, and
spermatozoa undulating their flagellae in low-Reynolds-number
environments, all of which are studied in biologic, physics, or
robotics contexts.

Overall, inner dissipation can model metabolic expenditures in biological systems or actuation effort in motorized robotic joints. 
Bending actuations had previously been included in the kinetic energy
for the inertia dominated setting~\cite{Hatton:2022:GOG,Yang:2023:GGO}.
With our combined outer and
inner dissipation we determine optimal motion trajectories using a variational principle.
In what follows we consider, for concreteness, shape changing creatures that
can be abstracted to one dimensional curves. 

To formulate optimal motion paths in the presence of dissipation mathematically,
we denote by \(\mathcal{M}\) the infinite dimensional manifold of all (smooth) positioned curves \(\gamma\). This manifold composes of \emph{fibers} consisting of all positionings---placement and orientation---of a given shape. Both the inner and outer dissipation are given as scalar products \(\langle \cdot,\cdot\rangle_{\rm in/out}\) on the tangent bundle \(T\cM\). The total dissipation of a deforming shape's motion trajectory---a path
\(\boldsymbol\gamma =(\boldsymbol\gamma_t)_{t\in [0,1]}\) in \(\mathcal{M}\)---is
now given as
\[
\cE(\boldsymbol\gam) \coloneqq \int_0^1 \langle {\boldsymbol\gam}'_t, {\boldsymbol\gam}'_t\rangle_\inn+ \langle {\boldsymbol\gam}'_t, {\boldsymbol\gam}'_t\rangle_\ext\, \dt.
\]
As a consequence of Helmholtz' principle of least dissipation~\cite{Helmholtz:1869:TSS} the motion trajectory \(\boldsymbol\gam\) induced by the shape variations 
\(\boldsymbol\gam'\) should be orthogonal to the fibers in \(\mathcal{M}\). We can compute  optimal motion paths as minimizers of the total dissipation subject to constraints and boundary conditions for \(\boldsymbol\gamma(0)\) and \(\boldsymbol\gamma(1)\), such as their positioning, their shape or periodicity.
Our constrained optimization model naturally embeds into sub-Riemannian calculus and the resulting paths are sub-Riemannian geodesics~\cite{montgomery:2002:ATo}.
To examine these comprehensively requires robust and efficient numerical tools able to handle high-dimensional optimization problems~\cite{Yang:2024:TGM}. 
While there are a variety of approaches that rely on, \eg, reinforcement learning~\cite{Jiao:2021:LTS, Qin:2023:RLM} we restrict the discussion to geometric approaches, closer to our own.

The predominant approach in the literature exploits the symmetries of the configuration space \(\cM\) under rigid body motions. 
This reduces optimal gait problems to optimization problems on the shape space \(\cS\) given by the quotient of the configuration space and the group of rigid body transformations~\cite{Tam:2007:OSP, Hatton:2012:KKL, Hatton:2017:KCE, Ramasamy:2019:GOG, Hatton:2022:GOG, Yang:2023:GGO}. 

%%%%%
Cyclic shape changes can be characterized by \emph{constraint curvature} 
and approximated using \emph{corrected velocity based integrals}~\cite{Hatton:2015:NNL}. 
This approach allows for low-dimensional representations of the gaits in terms of, \eg, Fourier series or splines
but is restricted to the treatment of closed gaits and relies on explicit shape space parameterizations~\cite{Hirose:2009:SLR}.
The quality of the resulting motion depends heavily on the choice of coordinates, which in practice necessitates additional optimization steps to determine optimal coordinate systems~\cite{Bass:2022:CEN}. 
%%%%%

Other methods
use explicit differentiation through the forward simulation%based on variational integration
~\cite{Becker:2025:IGL} and thereby enable the efficient approximation of solutions to sub-Riemannian boundary value problems. 
Though, the need for expressive and sufficiently regular shape spaces remains~\cite{Hirose:2009:SLR, Becker:2025:IGL}. 

Our variational characterization of the dynamics establishes a continuous model for optimal locomotion
accounting for both, outer and inner dissipation. We allow a variety of boundary conditions based on positioned shape, periodicity, or displacement restrictions.
Our approach recovers findings obtained with previous models used to simulate 
low-dimensional swimmers, such as Purcell's swimmer, and leads to energetically
more efficient motion paths.
We solve the resulting optimization problems efficiently 
allowing for shape changing locomotors at arbitrarily high resolution.
The method is applicable to a wide range of slender locomotors and compares favourably with experimental data.

\subsection{Outline}
The paper is organized as follows. 
In \secref{sec:AGeometricFrameworkForLocomotion} 
we detail the Riemannian description of locomotion and the fiber structure of configuration space, 
which hinges on a careful distinction between shape and placement in world space. 
In \secref{sec:AModelForEfficientGeometricLocomotion} we discuss outer and inner dissipation and 
explicate the different boundary conditions.
Finally, in \secref{sec:Results} we describe the numerical implementation, 
demonstrate the effectiveness of our approach in applications and compare to reduced models.

\section{A geometric framework for locomotion}
\label{sec:AGeometricFrameworkForLocomotion}
We begin our exposition by recalling some relevant concepts from differential geometry and introducing the necessary ideas to develop a geometric framework for locomotion. For concreteness' sake we consider the configuration space \(\cM\) of one-dimensional shapes in \(\RR^3\).  These form an infinite dimensional manifold that can be represented as the space of smooth regular curves positioned in $\RR^3$ 
\begin{equation*}
	\cM = \{ \gam \in C^\infty([0,1];\RR^3) ,\; s\mapsto \gam(s) \mid \partial_s \gam \neq 0\}.
\end{equation*}

Any \(\gam\in\cM\) in this configuration space of \emph{positioned
	shapes} can be re-positioned by the action of an orientation
preserving rigid-body motion \(g\in G \coloneqq \SE(3)\) without changing its shape. That is, the group \(G\) acts on the configuration manifold \(\cM\) by the action
\begin{equation*}
	G\times\cM\to\cM,\ (g,\gam)\mapsto g(\gam) = A\gam + b \in\cM,
\end{equation*}
where \(A\in\SO(3)\) (a rotation) and \(b\in\RR^3\) (a translation). Since rigid body transformations do not change the shape of \(\gam\), considering all points of the configuration space \(\cM\) that share the same shape induces a natural foliation of \(\cM\) into \(\dim(G)\)-dimensional submanifolds called \emph{fibers} 
\begin{equation*}
	G(\gamma) = \{ g(\gamma) \mid g \in G \}.
\end{equation*} 
Each of which can be identified with the group \(G\) itself. Hence, the
\emph{shape space}, is the quotient \(\cS = \cM/G\), equipped with the
canonical projection map \(\pi\colon\cM\to\cS\). That is, a
\emph{shape} is an element of \(\cM\) up to its position and orientation. 

\begin{figure}%[h]
	\centering
	\includegraphics[width = 0.8\linewidth]{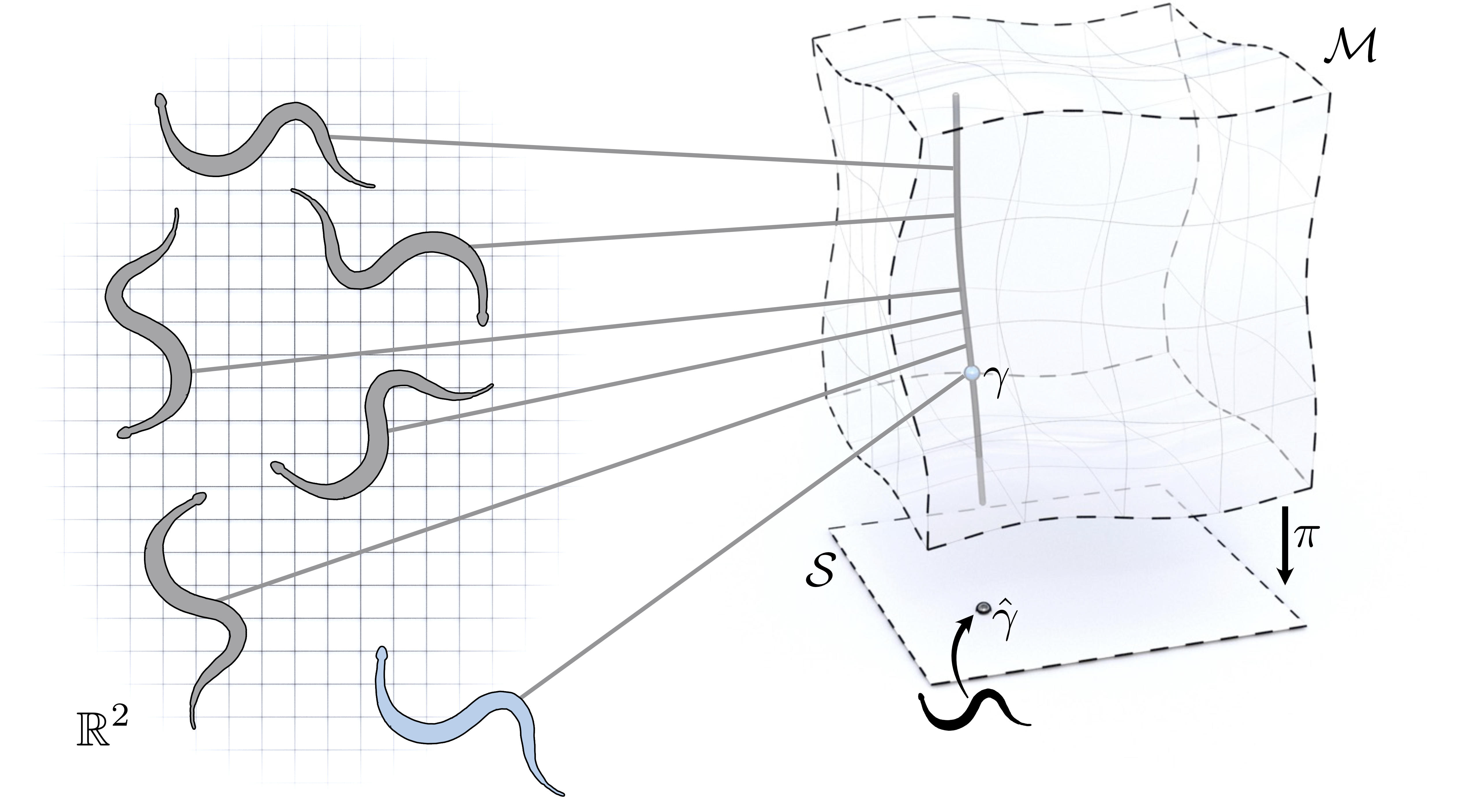}
	\caption{A shape (snake) can be variedly positioned
		(left). All of these configurations---being different only
		by a orientation preserving Euclidean motion---belong to
		the same fiber (right) and project down to a single shape in
		\(\cS\).}
	\label{fig:ShapeSpaceVsConfigrationSpace}
\end{figure}
In differential geometric terms, the fibered structure of \(\cM\) is a
\emph{principal fiber bundle}\footnote{For in depth discussions of
	geometric concepts such as fiber bundles, Lie groups, their geometry
	and application in physics, we refer to excellent and comprehensive
	references such as~\cite{Marsden:1999:IMS},
	\cite{montgomery:2002:ATo} or~\cite{Frankel:2011:GOP}.} over the
shape space \(\cS\) as the base space (\figref{fig:ShapeSpaceVsConfigrationSpace}). Moreover, at each configuration
\(\gam\in\cM\) the group action determines a special
subspace \[V_\gam\coloneqq\ker(d\pi_\gam)\subset T_\gam\cM\] of all
vectors tangent to the six-dimensional fiber \(G(\gam)\) at
\(\gam\). Vectors in this \emph{vertical subspace} are in a one-to-one
correspondence with infinitesimal rigid body transformations. 
The collection of all vertical subspaces \(V_\gam\) 
defines the \emph{vertical subspace field} \(V\).

\subsection{Shape deformations and motion trajectories}
Leaving motion aside, one may ask for an optimal deformation from
one shape into another. Here, \emph{optimal} can, for example, be the
shortest path between two points in the shape space $\cS$. 
Within the above setup, the smooth deformations of a shape changing body---modulo its
positioning in world space---correspond to a smooth one-parameter
family \(\boldsymbol{\hgam}=(\boldsymbol{\hgam}_t)_{t\in [0,1]}\). 
Each \(\boldsymbol{\hgam}_t\colon[0,1]\to\cS\) describes the shape at time
\(t\in[0,1]\). Finding such optimal paths requires a metric. A
classic approach uses Rayleigh's analogy~\cite{Rayleigh:1871:TRV} 
which measures \emph{inner dissipation}, \ie, the energy needed to affect the implied
elastic deformation of a shape \(\boldsymbol{\hgam}_t\) (see \secref{sec:InnerDissipation} for more details). Note that
in this setup rigid motions are excluded.

A motion trajectory of a shape changing body describing its
position and orientation in world space corresponds to a smooth
one-parameter family \(\boldsymbol{\gam}=(\boldsymbol{\gam}_t)_{t\in [0,1]}\) with
\(\boldsymbol{\gam}_t\colon [0,1]\to \cM\). A motion path \(\boldsymbol\gamma\in \cM\) is called a \emph{lift} of \(\hat{\boldsymbol{\gam}}\) if its projection matches the given shape sequence
\(\pi(\boldsymbol{\gam}_t)=\boldsymbol{\hgam}_t\) for all \(t\in[0,1]\). Yet again, it is natural to ask for \emph{optimal} motion paths, this time with
respect to a metric on \(\cM\). 
Besides the inner dissipation this metric accounts for \emph{outer dissipation}, 
\ie, the friction due to motion induced interaction with the environment (see \secref{sec:outerdissipation} for more details).
It is a goal of the present paper to establish such a metric that accounts for both types of dissipation.
\begin{figure}[h]
	\centering
	\includegraphics[width = 0.8\linewidth]{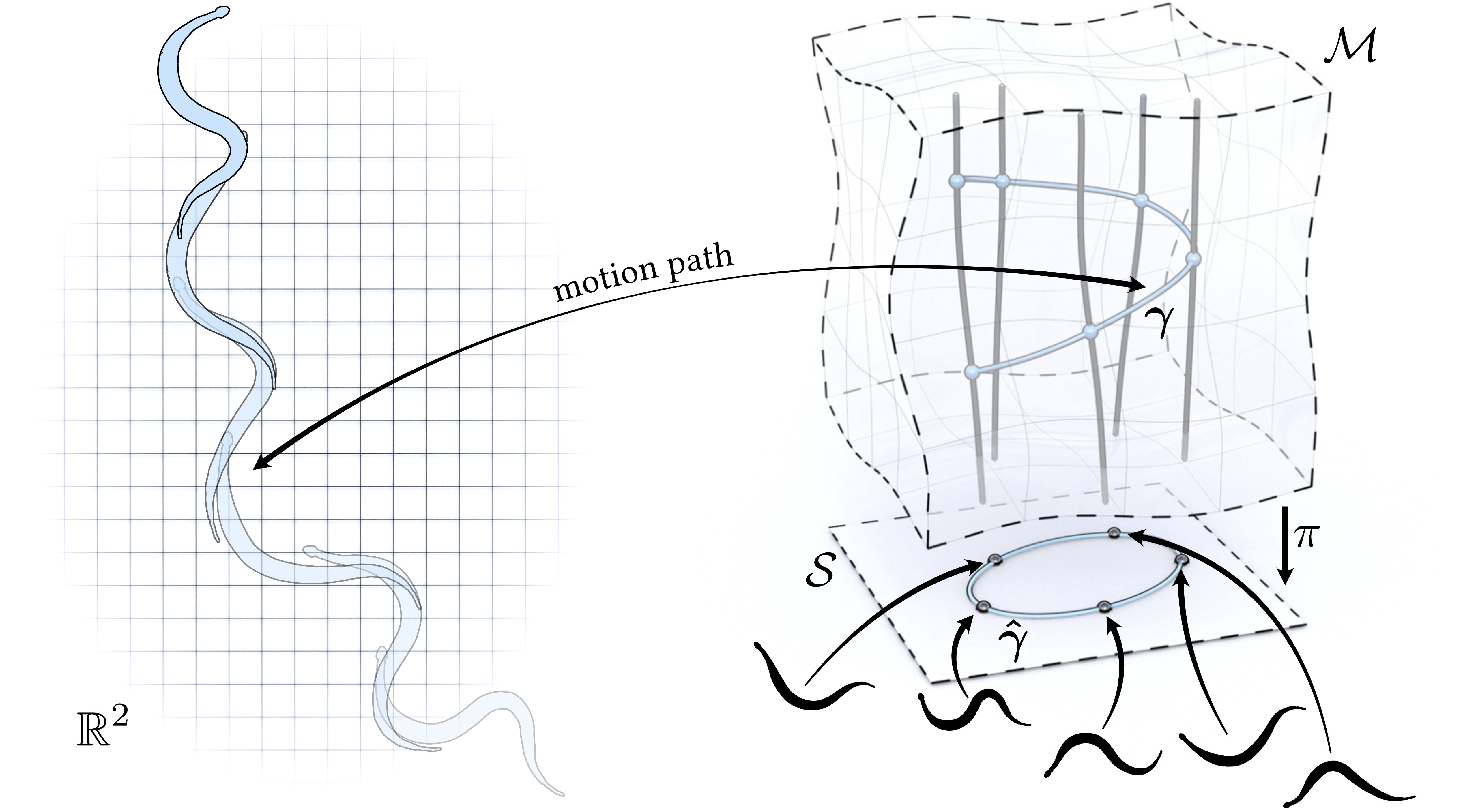}
	\caption{A cylic motion of a changing shape (left) amounts to a
		closed curve in \(\cS\) under the projection \(\pi\) (right). The
		fact that the cyclic shape deformation results in net motion
		per cycle is seen in \(\boldsymbol{\gamma}\) beginning and
		ending in different locations on the same fiber (right).}
	\label{fig:GeometricLocomotion}
\end{figure}

\subsection{Physical motion paths} \label{sec:PhysicalMotionPaths}
It is now natural to ask: what distinguishes
the motion paths that are actually observed in nature? The present
investigations focus on scenarios of \emph{geometric locomotion},
where motion is entirely governed by a set of conserved quantities. It
encompasses the dynamics of systems ranging from astronauts
maneuvering in zero gravity, to shape-changing microorganisms, to the
terrestrial locomotion of snakes. More generally, it applies whenever
the physical motion in $\cM$ is fully determined by the path in shape
space \(\cS\)~\cite{montgomery:2002:ATo, Hatton:2012:KKL,
	Gro23MfSC,Glisman:2022:SPF}. 

We adopt a Lagrangian point of view to describe the locomotion of a shape-changing body as a path \(\boldsymbol{\gam}\) in \(\cM\). Then, physical motions obey principles of least action, \ie, are stationary points of the path energy
\begin{equation}
	\label{eq:GenericQuadraticForm}
	\cE(\boldsymbol\gamma)=\tfrac{1}{2}\int_0^1\langle \boldsymbol\gam'_t, \boldsymbol\gam'_t\rangle_\cM\,\dt,
\end{equation}
under constraints on the velocity vectors \(\boldsymbol\gam'_t \in T_{\gam_t}\cM\) of the curve \(\boldsymbol{\gam}\) in \(\cM\)
at time $t$. 

The geometry of the configuration space of positioned shapes---and hence the modeled physics---is determined by the choice of a Riemannian metric \(\langle \cdot, \cdot \rangle_\cM\) on the configuration manifold \(\cM\). To respect the principal fiber structure of \(\cM\), this metric must be invariant under the action of \(G\). That is, for all \(\gam \in \cM\), \(X, Y \in T_\gam \cM\), and \(g \in G\),
\begin{equation}
	\label{eq:GInvarianceMetric}
	\langle X, Y \rangle_{\cM} = \langle g(X), g(Y) \rangle_{\cM}.
\end{equation}
With the \emph{\(G\)-invariance} of the metric \(\langle \cdot, \cdot \rangle_\cM\), also \teqref{eq:GenericQuadraticForm} becomes invariant under global rigid body transformations, \ie, \(\cE(g \circ \boldsymbol{\gamma}) = \cE(\boldsymbol{\gamma})\) for all \(g \in G\). That is, applying any global rigid-body transformation to the entire trajectory leaves its energy (\teqref{eq:GenericQuadraticForm})  unchanged. Typical examples for the choice of a metric include the kinetic energy, thus modeling inertia-dominated motion in the absence of forces~\cite{Kulwicki:1962:WM}, or the energy dissipation caused by the body's displacement in an ambient medium, which models the locomotion of organisms in both low-Reynolds number~\cite{Hu:2009:MSL} and high Reynolds number environments~\cite{Glisman:2022:SPF}.

By Noether's theorem~\cite{Noether:1918:IVP} the symmetry
(\teqref{eq:GInvarianceMetric}) implies the existence of six conserved
quantities, one for each dimension of the infinitesimal rigid body
motions.  This will allow for the reconstruction of
motion trajectories $\boldsymbol\gamma$ 
for given paths \(\boldsymbol{\hat\gamma}=(\boldsymbol\hgam_t)_{t\in[0,1]}\) in $\cS$ (\figref{fig:GeometricLocomotion}),
which are the unique lift (up to a global rigid body motion) of \(\boldsymbol{\hat\gamma}\)
that is a stationary point of the energy
(\teqref{eq:GenericQuadraticForm}).

Here, a fundamental result  is that a smooth path $\boldsymbol{\gam}$ in $\mathcal M$ is a stationary point of the energy
under temporally smooth vertical variations 
$\boldsymbol{Z} = (\boldsymbol{Z}_t)_{t\in[0,1]} \in V_{\boldsymbol{\gam}}$
if and only if, the horizontality constraint 
\begin{equation}
	\langle\boldsymbol{\gam}'_t,\boldsymbol{Z}_t\rangle_\cM = 0 
	\label{eq:vertical-orthogonality} 
\end{equation}
holds  for all $t \in [0,1]$ and for all variations 
$\boldsymbol{Z}\in V_{\boldsymbol{\gam}}$ \cf\cite{Gallot:1990:RG, Gro23MfSC}. 
For the sake of completeness we provide a derivation of this constraint in (\cf\@ \cref{supp:horizconstraint}).
In other words, the six degrees of freedom
associated with the positioning of each shape in space are constrained by six
independent conditions and a motion path is physical if and only if the vertical component of \({\boldsymbol\gam}'\) vanishes
along the trajectory. 
must hold for any lift, it follows that any physical motion path
must traverse the fibers of \(\cM\) orthogonally with respect to the Riemannian metric. This special role of the orthogonal complement justifies the definition of the \emph{horizontal subspace}
\begin{equation*}
	\cH_\gam\coloneqq 
	\{X\in T_\gam\cM\mid \langle X,Z\rangle_\cM = 0,\ \forall Z\in V_\gamma \}\subset T_\gam\cM
\end{equation*}
and the associated horizontal subspace field \(\cH \) as the space of all \( \cH_\gam\) for \(\gam\in\cM\).
Any vector \(X\in T_\gam\cM\) can be uniquely decomposed into horizontal and vertical components 
\begin{equation*}
	X=X^V + X^\cH,
\end{equation*}
where \(X^V\in V_\gam\) and \(X^\cH\in\cH_\gam\) (\cf~\cite[Ch.\,11]{montgomery:2002:ATo}). Consequently, motion paths for which \(\boldsymbol\gam'\in\cH_{\boldsymbol{\gam}}\) are said to be  \emph{horizontal} and \eqref{eq:vertical-orthogonality} states that the stationary points of \teqref{eq:GenericQuadraticForm} under vertical variations are given by \emph{lifts} which are \emph{horizontal}. 

%%%%%%%%%%%%%%%%%%%%%%%%%%%%%%%%%%%%%%%%
%%%%%%%%%%%%%%%%%%%%%%%%%%%%%%%%%%%%%%%%
%%%%%%%%%%%%%%%%%%%%%%%%%%%%%%%%%%%%%%%%
\section{A model for energy efficient geometric locomotion}
\label{sec:AModelForEfficientGeometricLocomotion}

Having established the characterization of physical motion paths as \emph{horizontal lifts}, we now turn to our primary objective, a geometric model for energy efficient locomotion paths. 
Note that, although our proposed model is applicable to various
settings~\cite{montgomery:2002:ATo, Hatton:2012:KKL, Glisman:2022:SPF,  Gro23MfSC}, 
we focus on dissipation dominated scenarios. 
Thus, energy efficient locomotion paths are paths $\boldsymbol{\gam} \subset \cM$ which are
admissible in the sense that they result from horizontal lifts, \ie, $\boldsymbol{\gam}' \in \cH_{\boldsymbol{\gam}}$ and minimize dissipation. 
Hence, optimal paths are obtained as solutions to 
\begin{equation}
	\label{eq:InitialMentionOfArgminProblem}
	\underset{\boldsymbol\gam \in\cA_{\cH}}{\argmin}\,  \cE( \boldsymbol{\gam}).
\end{equation}
where $\cA_{\cH}$ is the set of admissible motion paths $\boldsymbol{\gam}$, which fulfill 
$\boldsymbol{\gam}' \in \cH_{\boldsymbol{\gam}}$ and obey application dependent boundary conditions to be discussed below in \suppsecref{sec:OptimalLocomotionProblems}. 

To measure dissipation, we propose a family of \(G\)-invariant Riemannian
metrics on the configuration space \(\cM\). These metrics capture 
locomotion efficiency in a purely geometric fashion. This establishes a novel
geometric mechanics model of shape change induced motion, which
offers a more comprehensive understanding of energy efficient
locomotion.

%%%%%%%%%%%%%%%%%%%%%%%%%%%%%%%%%%%%%%%%
\subsection{Dissipation metrics}
\label{sec:DissipationMetrics}

Our dissipation model combines an \emph{outer dissipation} 
modeling the interactions between the body and the environment,
and an  \emph{inner dissipation} accounting for the energetic cost of
bending and straining (stretching/compression) of the body.

%%%%%%%%%%%%%%%%%%%
\subsubsection{Outer dissipation}\label{sec:outerdissipation}
\begin{figure}
		\centering
		\includegraphics[width = .8\linewidth]{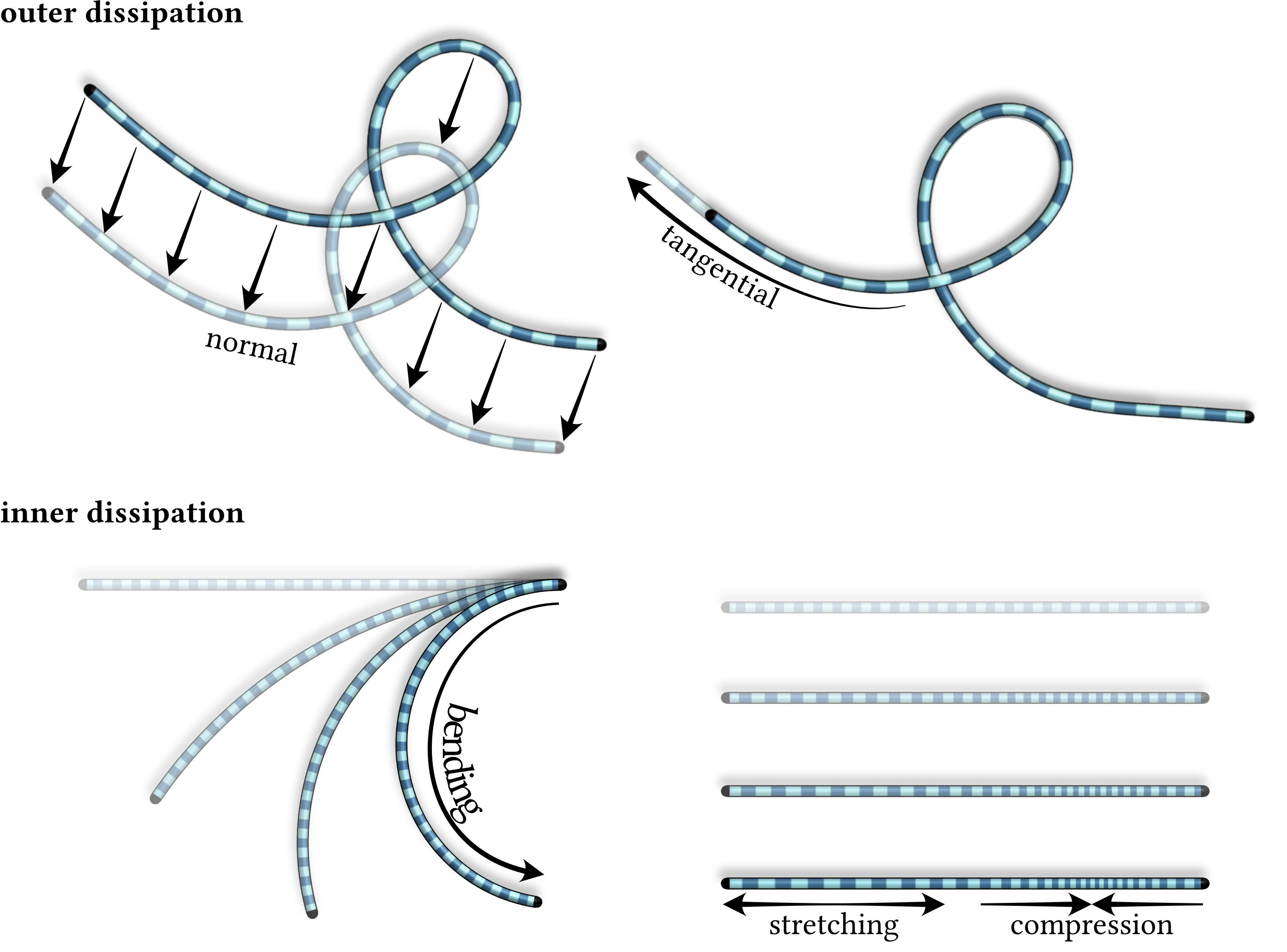}
		\caption{\figloc{Top:} Outer dissipation accounts for energy expenditure due to external forces, such as viscous friction. In general we expect normal motion (left) to be
			more expensive while tangential motion (right) is less so. This
			anisotropy, which may vary along the length of the
			creature is responsible for deformations
			resulting in net motion.
			\figloc{Bottom:} Inner dissipation accounts for energy spent on internal shape changes and results from metabolism 
			or from sources such as batteries. First order
			effects correspond to stretching and compression while
			bending is a second order phenomenon.
		}
		\label{fig:InnerVsOuter}
	\end{figure}

Displacements of a body immersed in, \eg, a highly viscous medium dissipate energy  due to friction. 
To model such viscous friction effects, we adopt the framework of Rayleigh dissipation.
The rate of dissipation caused by infinitesimal
configuration changes of the body is, in general, anisotropic and
varies along the shape. For instance, a slender body like a cylinder
experiences less drag when displaced along its longitudinal axis compared to
displacements in transverse directions
(\figref{fig:InnerVsOuter}, top). For shape-changing bodies, like a bacterium swimming with flagellar motion, the situation
becomes more intricate, with preferred directions being both local and time-dependent quantities. 
To accommodate these anisotropies in our
model, we introduce a map
\begin{equation}
	\label{eq:BMap}
	\cB\colon\cM\to C^\infty([0,1];\RR^{3\times 3})
\end{equation}
To each configuration \(\gam\in\cM\) it associates a map
\(\cB_\gam\) with values in the symmetric positive definite matrices
and representing an anisotropic rescaling of infinitesimal configuration changes.
We suppose that $\cB_\gam$ is invariant under the group action on \(\cM\), \ie,
\(g(\cB_\gam) = \cB_{g(\gam)}\) for all \(\gam\in\cM\) and \(g\in G\). 

The instantaneous dissipation caused by the infinitesimal shape
change \(X_\gam\in T_\gam\cM\) is then obtained by integrating the local
dissipation density
\(\langle \cB_\gam X_\gam,X_\gam\rangle_{\RR^3}\) over the whole
body.  This leads to the \emph{outer dissipation metric} as a 
non-degenerate and \(G\)-invariant inner product on \(\cM\).
\smallskip

\subsubsection*{Definition of outer dissipation} 
{\it 
	For \(X\in T_\gam\cM\), the \emph{outer dissipation} metric at the body $\gam$ is defined by 
	\begin{equation}
		\label{eq:OuterDissipationDef}
		\langle X,X\rangle_\ext = \tfrac{1}{2}\int_\gam \langle \cB_{\gam} X_{\gam},X_{\gam}\rangle_{\RR^3}\,.
	\end{equation}
}

If the body is moving in a viscous Stokes flow the evaluation of the metric tensor \(\cB\) 
requires the solution of an elliptic boundary value problem.
However, in practice it is often more tractable to approximate outer 
dissipation using \emph{resistive force theory}~\cite{Gray:1955:PSU,Zhang:2014:TEO}. 
It neglects (rapidly decreasing~\cite{Chwang:1975:HLR}) long-range interactions and defines 
for an anisotropy ratio $\epsilon  \in (0,1]$
\begin{equation}\label{eq:Bgamma}
	\cB_\gam = \id + (\epsilon-1) P_{\gam},
\end{equation}
where \(P_{\gam}\) denotes the projection onto the tangential space of the body.
This implies isotropic dissipation for  \(\epsilon =1\), while for \(\epsilon\approx 0\), tangential motion incurs negligible dissipation
(\cf\@ \cref{suppsec:DiscreteLagrangian})).

%%%%%%%%%%%%%%%%%%%
\subsubsection{Inner dissipation}
\label{sec:InnerDissipation}
In contrast to outer dissipation, inner dissipation arises from the
activation of a body's intrinsic degrees of freedom. 
Our model accounts for internal material stresses caused either by bending or straining. 
We use them to model   
the energy expended during the locomotor's deformation, such as the
energy consumed by an organism's metabolism or the battery power
used to drive robot actuators
(\figref{fig:InnerVsOuter}, bottom).
We consider variations $\delta\gam \in T_\gam\cM$ of positioned shapes $\gam$. These induce variations 
\(\delta\kappa\) as instantaneous changes of the body's curvature and
variations  $\delta \vert \partial_s\gam\vert$ measuring variations of the length element, 
\ie, the stretching or compression of the body 
(\cf\@\cref{suppsec:DiscreteLagrangian} for a detailed treatment).

We consider the weighted sum  
\begin{equation}
	\label{eq:InnerDissipationW}
	\cW_{\gam}[\delta\gam] = \wben\cWben[\delta\gam] + \wmem\cWmem[\delta\gam]
\end{equation}
of the two nonlinear energies
\begin{equation}\label{eq:WbenWmem}
	\cWben = \int_{\gam}(\delta\kappa)^2 \textit{ and } 
	\cWmem = 
	\int_\gam\left\vert \frac{ \delta\left\vert\partial_s\gam\right\vert}{\left\vert \partial_s\gam\right\vert} \right\vert^2 
\end{equation}
measuring dissipation caused by bending \resp\@ strain distortion induced by the variation  $\delta\gam$
~\cite{bergou2008discrete,  Mio:2007:OSS, Bauer:2019:ARA}.
The linearization of $\cW$ at a fixed time and a fixed body $\gam$ leads to the quadratic form on 
$X = \delta\gam \in T_\gam\cM$ induced by the Hessian $\Hess \cW_\gam[0]$ and gives rise to the definition of the inner dissipation.

\subsubsection*{Definition of inner dissipation}
{\it 
	For \(X\in T_\gam\cM\) for a given body $\gam$ the \emph{inner dissipation} is defined as the symmetric quadratic form 
	\begin{equation}
		\label{eq:InnerDissipation}
		\langle X,X\rangle_\inn\coloneqq \tfrac{1}{2}\int_{\gamma}\Hess \cW_\gam[0](X_\gam,X_\gam)\,.
	\end{equation}
}

As \(\cW\) is \(G\)-invariant, so is the quadratic form \(\langle \cdot,\cdot\rangle_\inn\), \ie,  constant along the fibers. 
The Hessian evaluated at the identity is symmetric semi positive-definite with a six-dimensional kernel reflecting the infinitesimal rigid body motions. Therefore, it is degenerate in directions of vertical vector fields, and does not on its own constitute a metric on the configuration space. However, it is non-degenerate when restricted to the horizontal subspace field.
Examples for other quadratic forms measuring the quality of motion paths can in general be obtained by pulling back any Riemannian metric \(\langle \cdot,\cdot\rangle_\cS\) from the shape space to the configuration space, \ie, by setting 
\begin{equation*}
	\langle \cdot,\cdot\rangle_\inn \coloneqq \langle d\pi(\cdot),d\pi(\cdot)\rangle_\cS.
\end{equation*}

\subsubsection{Total dissipation}
\label{sec:TotalDissipationMetric}
Combining \(\langle \cdot,\cdot\rangle_\inn\) and \(\langle \cdot,\cdot\rangle_\ext\) we obtain a non-degenerate Riemannian metric 
on the configuration space \(\cM\) that captures the effects of both, inner and outer dissipation. 

The \emph{total dissipation} metric at a body $\gam$ is given by 
\begin{equation}
	\label{eq:TotalDissipationMetric}
	\langle\cdot,\cdot\rangle_\cM = \langle\cdot,\cdot\rangle_\inn + \langle\cdot,\cdot\rangle_\ext.
\end{equation}

The total dissipation metric inherits the compatibility with the principal fiber bundle structure of the configuration space \(\cM\).

%%%%%%%%%%%%%%%%%%
%%%%%%%%%%%%%%%%%%
%%%%%%%%%%%%%%%%%%
\subsection{Different optimal locomotion problems}
\label{sec:OptimalLocomotionProblems}
We now turn to the geometric formulation of optimal
locomotion problems with application dependent boundary conditions. 
Rather than lifting a prescribed shape sequence
to recover the resulting motion, we directly seek optimal motion sequences of 
shape-changing bodies in world space minimizing total dissipation. 

%%%%%
The motion of snakes, worms, soft robots, or ice skating
involve \emph{non-holonomic} constraints for the motion velocity. 
As described in \secref{sec:PhysicalMotionPaths}
physical motion paths are inherently confined to the
horizontal subspace field. Consequently optimal locomotion strategies
must respect this constraint (\teqref{eq:InitialMentionOfArgminProblem}). This leads naturally to the framework of \emph{sub-Riemannian geometry}, which provides a rigorous and practical language for formulating and analyzing problems of constrained optimal motion~\cite{montgomery:2002:ATo}.
In our context the triple $(\cM,\cH,\langle\cdot,\cdot\rangle^\cH_\cM)$ defines a particular case of 
a sub-Riemannian manifold, where the metric \(\langle\cdot,\cdot\rangle^\cH_\cM\) is defined as the 
restriction of the Riemannian metric \(\langle\cdot,\cdot\rangle_\cM\)
to the horizontal subspace field \(\cH\), called \emph{Carnot-Caratheodory metric}.
Horizontal paths that minimize the energy with respect to this metric are called \emph{sub-Riemannian geodesics} and exhibit constant velocities~\cite[App.D]{montgomery:2002:ATo}.

Remarkably, the dissipation of a motion path \(\boldsymbol\gam\) is
fully captured by \(\langle\cdot,\cdot\rangle^\cH_\cM\), 
since admissible motion paths considered in the optimization problem (\teqref{eq:InitialMentionOfArgminProblem})
must be tangent to the horizontal subspace field. 
Thus, replacing the quadratic form $\langle\cdot,\cdot\rangle_\cM$ in (\teqref{eq:InitialMentionOfArgminProblem})
by the restricted quadratic form $\langle\cdot,\cdot\rangle^\cH_\cM$ does not effect minimizers.
Yet, it is the vertical subspace field that determines the dynamics of a shape changing body.

%%%%%%%
In what follows, we introduce three optimal locomotion problems, which are
distinguished by differing constraints on the initial and target
shapes, as well as their respective start and end positions.

%%%%%%%%%%%%%%%%%%%%%%%%%%%%%%%%%%%
\subsubsection{Fixed initial and target body in world coordinates}
\label{sec:Dirichlet}
The most direct boundary condition fixes the initial body $\gam^0$ at time $0$
and the target body $\gam^1$ at time $1$. Thus, we minimize the total dissipation $\cE( \boldsymbol{\gam})$ over 
the set 
\begin{equation*}
	\cA^{\text{fix}}_{\cH}\coloneqq\{\boldsymbol{\gam}\colon[0,1]
	\to\cM\mid \boldsymbol{\gam}_0=\gam^0,\, \boldsymbol{\gam}_1=\gam^1,\, \boldsymbol{\gam}'\in\cH_{\boldsymbol{\gam}}\}
\end{equation*}
of admissible paths $\boldsymbol{\gam}$.
As mentioned earlier \emph{sub-Riemannian geodesics}, \ie, length minimizing paths with respect to the metric
$\langle\cdot,\cdot\rangle^\cH_\cM$ in $\cA^{\text{fix}}_{\cH}$ coincide with solutions of 
\begin{equation}
	\label{eq:geodesic}
	\underset{\boldsymbol\gam \in\cA^{\text{fix}}_{\cH}}{\argmin}\,  \cE( \boldsymbol{\gam}).
\end{equation}
Based on the so-called \emph{H\"ormander's condition} the Chow-Rashevskii theorem assures that 
\(\cA^{\text{fix}}_{\cH}\) is non-empty for any pair of configurations \(\gam^0,\gam^1 \in \mathcal{M}\). This is also referred to as \emph{strong controllability}~\cite{montgomery:2002:ATo} 
and is the essential ingredient for the existence of an optimal physical motion path
between any two positioned shapes that solves \eqref{eq:InitialMentionOfArgminProblem}. 
Rigorous results on the strong controllability have previously been established for a variety of systems, including those discussed in this paper~\cite{Kadam:2016:CPS, Giraldi:2013:COS, Moreau:2023:COC}. 

\begin{figure}[h]
	\centering
	\includegraphics[width = 0.8\linewidth]{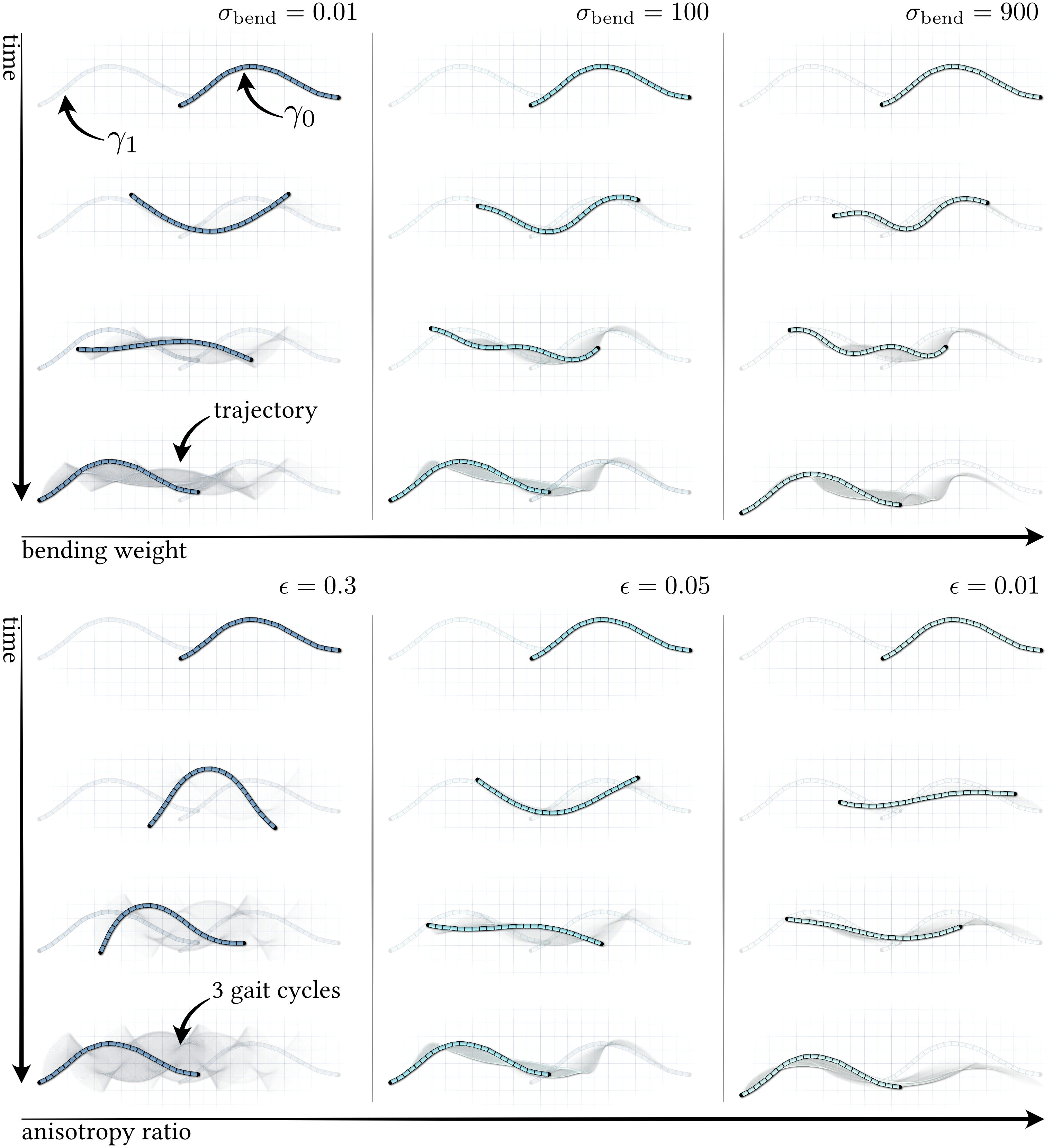}
	\caption{Solution of the boundary value problem with prescribed initial and target body in 
		world coordinates (\teqref{eq:geodesic})  for different bending weights  
		and fixed strain weight $1.0$ controlling the inner dissipation (top), for different anisotropy ratios controlling the outer dissipation (bottom).}
	\label{fig:WeightsComparison}
\end{figure}

For our model example of a snake slithering on a plane, a solution to the sub-Riemannian boundary value problem in \teqref{eq:geodesic} answers the question how it most efficiently slithers from a given initial position and pose to a desired target position and pose (\figref{fig:BdyValuesProblems} top).
The impact of the different weighting of the terms of the inner dissipation 
and the anisotropy ratio in the outer dissipation are shown in \figref{fig:WeightsComparison}.

%%%%%%%%%%%%%%%%%%%%%
%%%%%%%%%%%%%%%%%%%%%

%%%%%%%%%%%%%%%%%%%%%%%%%%%%%%%%%%%%%%%%%%%%%%%%%
%%%%%%%%%%%%%%%%%%%%%%%%%%%%%%%%%%%%%%%%%%%%%%%%%
\subsubsection{Periodic boundary conditions} \label{subsec:periodicBoundaryCond}
Applications in robotics, for instance, often only require the determination of shape sequences that are optimal in the sense of minimizing energy consumption while realizing a desired displacement of the body instead of prescribing shapes. These, so-called \emph{isoholonomic} problems can be categorized into those that seek to determine periodic shape sequences and those that impose constraints on the start shapes (see \secref{sec:mix}).

Cyclic shape changes in the shape space $\cS$ can produce global motion, such as translation or rotation, 
even when infinitesimal changes are confined to the horizontal subspace field, which is related to the geometric concept of
holonomy~\cite{Pinkall:2024:DG, Frankel:2011:GOP}. 

Such periodic shape sequences, are commonly referred to as \emph{gaits} and often regarded as fundamental building blocks for various tasks, such as robotic control. Each gait affects the displacement of a locomotor by some fixed rigid body transformation \(g\in G\).  Repeated, say \(k\)-fold, execution of a gait consequently yields a displacement by \(g^k\in G\). 
Additionally, stringing together sequences of gaits that result in different translations and/or rotations provides a straightforward approach to effective control. 
Thus, the task of finding optimal gaits amounts to finding an optimal, cyclic shape sequence that displaces the locomotor by a prescribed rigid body transformation \(g\in G\) while dissipating the least amount of energy. This rigid body transformation is called holonomy.
We denote the set of motion paths in \(\cM\) that are (1) tangent to the horizontal subspace field \(\cH\),
(2) correspond to some periodic shape sequence in \(\cS\), and (3) exhibit a given holonomy  \(g\) by 
\begin{align*}
	\cA^{\text{per}}_{\cH,g}
	\coloneqq\{{\boldsymbol\gam}\colon[0,1]\to\cM\; \mid \;
	&g(\boldsymbol{\gam}_0)=\boldsymbol{\gam}_1,\, \pi(\boldsymbol{\gam})\in C^\infty(S^1;\cS),\, \\
	& \boldsymbol{\gam}'\in\cH_{\boldsymbol{\gam}},\, \mathcal{L}(\boldsymbol{\gam}_0) > 0\}.
\end{align*}

\begin{figure}[h]
	\centering
	\includegraphics[width = 0.9\textwidth]{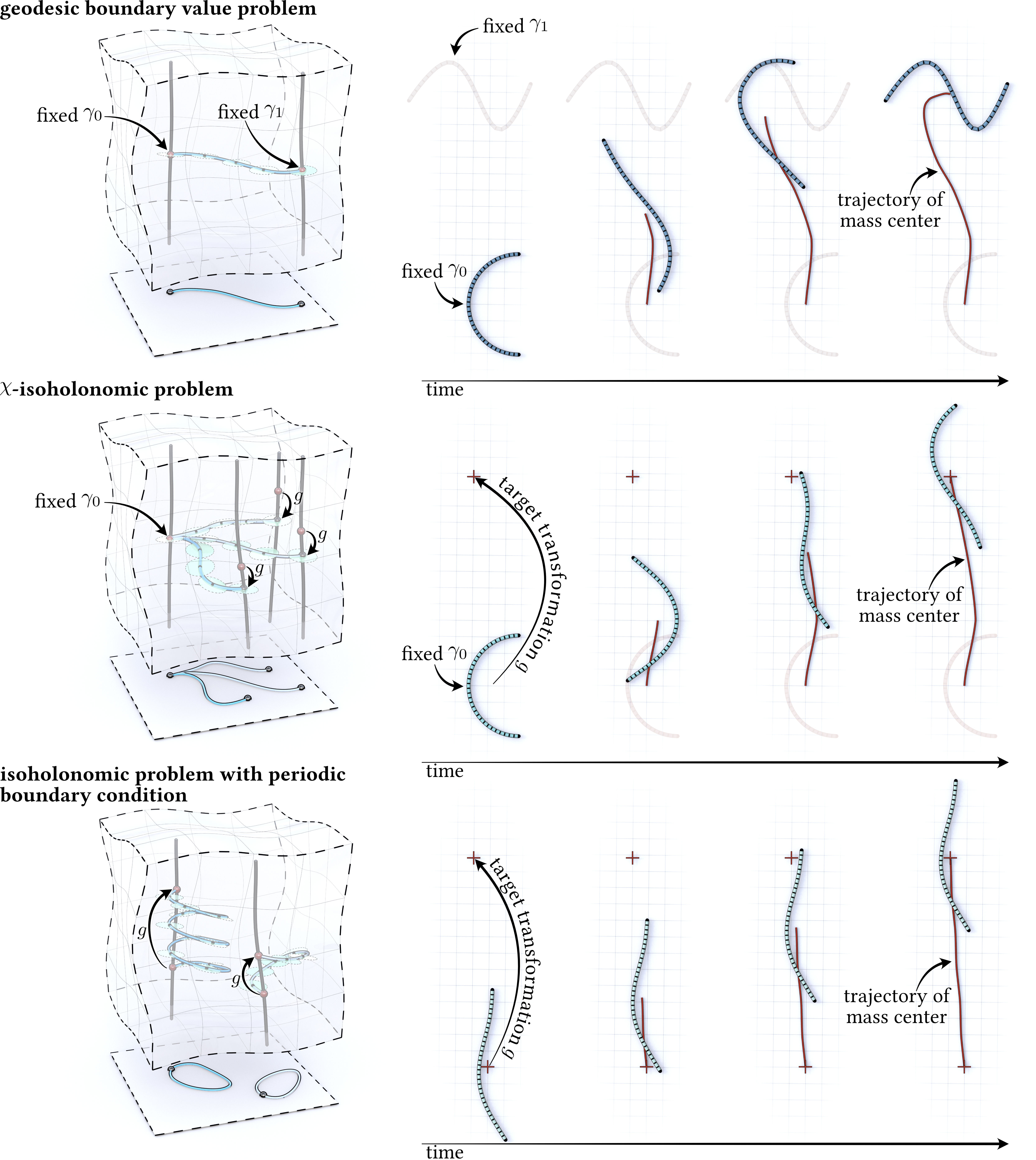}
	\caption{\figloc{Top:} optimal motion path between initial \(\gamma^0\)
		and final \(\gamma^1\) positioned shapes; \figloc{Middle:} fixing only the displacement
		from some initial configuration without requiring a
		particular end shape and thus with non prescribed destination fiber; 
		\figloc{Bottom:} optimal periodic motion path resulting in a rigid body motion (here a translation \(g\)) of the initial shape.             
		Each of the drawings in the left column visualizes the boundary constraints from a sub-Riemannian point of view.}
	\label{fig:BdyValuesProblems}
\end{figure}

Since our model allows length distortions of the locomotor, we have to explicitly rule out the constant path consisting of collapsed bodies 
for all times in $[0,1]$. This trivial solution can be avoided by the additional constraint that for one $t\in [0,1]$ the shape \(\boldsymbol{\gam}_t\) has strictly positive length \(\mathcal{L}(\gamma_t) > 0\) with
$\mathcal{L}(\gamma)= \int_\gamma \mathrm{d}s$.
Here, we required this constraint for $t=0$. 
We are led to the constrained optimization problem
\begin{equation}
	\label{eq:geodesicPeriodic}
	\underset{\boldsymbol\gam \in\cA^{\text{per}}_{\cH,g}}{\argmin}\,  \cE( \boldsymbol{\gam}).
\end{equation}
An example of optimal periodic motion is given in (\figref{fig:BdyValuesProblems} bottom).
%%%%%%%%%%
%%%%%%%%%%

%%%%%%%%%%%%%%%%%%%%%%%%%%%%%%%%%%%%%%%%%%%%%%%%%
%%%%%%%%%%%%%%%%%%%%%%%%%%%%%%%%%%%%%%%%%%%%%%%%%
\subsubsection{$\chi$-isoholonomic boundary conditions}\label{sec:mix}
Now,  we consider the following optimization task: 
given a snake-like robot with both its shape and position in world space fixed, 
we determine the optimal motion path that realizes a desired rigid body displacement without  constraining the final shape. 
This corresponds to  an \emph{isoholonomic} problem (see \secref{subsec:periodicBoundaryCond}) and the notion of \emph{weak controllability}~\cite{Kadam:2016:CPS}.
The difficulty is to define a reasonable matching rigid body motion for two bodies in world coordinates with different shape. 

For this one needs a canonical reference for shapes in world coordinates.
To address this, we introduce a \emph{reference gauge}
\begin{equation}\label{eq:chiRef}
	\chi\colon\cS\to\cM, \, \hat \gamma \mapsto \chi(\hat \gamma) \quad \text{such that}\ \pi\circ\chi=\id_\cS,
\end{equation}
where $\id_\cS$ is the identity map on the shape space $\cS$.
Intuitively, the reference gauge assigns a specific registration in world space to every shape. 

Any other configuration $\gamma$ in $\cM$ 
of the same shape $\hat \gamma=\pi(\gamma)$ in $\cS$ 
can then be uniquely expressed relative to 
\(\chi(\hat \gamma)\) 
as \(\gamma=g(\chi(\hat \gamma))\) for some \(g\in G\).

In practice, a reference gauge \(\chi\) can be constructed, \eg, by rigidly registering all shapes to a fixed reference configuration \(\gamma\in\cM\) (see, \eg,~\cite{Becker:2025:IGL}).

We consider a horizontal motion path \(\boldsymbol{\gam}\) in
\(\cM\) with a fixed gauge \(\chi\)~(\eqref{eq:chiRef}). 
Let \(g_0,g_1\in G\) such that \(\boldsymbol{\gam_0} = g_0(\chi(\boldsymbol{\hgam_0}))\) and
\(\boldsymbol{\gam}_1 = g_1(\chi(\boldsymbol{\hgam}_1))\). Then, by the \(G\)-invariance of the metric and the horizontal constraint, 
we may displace the whole motion path
\(\boldsymbol{\gam}\) by \(g_0^{-1}\) to start at
\(\chi(\boldsymbol{\hgam}_0)\in\cM\). In particular, this re-positioning lets
the displaced motion path end at the configuration
\(g_0^{-1}(\boldsymbol{\gam_1}) = g_0^{-1}g_1(\chi(\boldsymbol{\hgam_1}))\).
We call \(g_0^{-1}g_1\in G\)  the \(\chi\)\emph{-holonomy} of \(\boldsymbol{\gam}\)
and two horizontal motion paths with the same
\(\chi\)-holonomy are considered \(\chi\)\emph{-isoholonomic}. 
Assuming without loss of generality that 
$\boldsymbol{\gam}_0 = \chi(\boldsymbol{\hgam}_0)$ 
we obtain the set of admissible motion paths 
\begin{equation*}
	\cA^{\chi}_{\cH,g} 
	\coloneqq\{{\boldsymbol\gam}\colon[0,1]\to\cM\mid  \boldsymbol{\gam}_0=\gam^0,\, g(\gam^0)= \boldsymbol{\gam}_1,\, \boldsymbol{\gam}'\in\cH_{\boldsymbol{\gam}}\}
\end{equation*}
for a given \(\chi\)-holonomy \(g\) with reference gauge \(\chi\).
An optimal motion path starting at a configuration \(\gam^0\in\cM\) and realizing a given $\chi$-holonomy \(g\in G\) 
is a solution to the following \emph{\(\chi\)-isoholonomic} gait problem
\begin{equation} 
	\label{eq:ChiHolonomicProblem}
	\underset{\boldsymbol\gam\in\cA^{\chi}_{\cH,g}}{\argmin} \; \cE( \boldsymbol{\gam}).
\end{equation}
Also for this boundary condition an example is given in
(\figref{fig:BdyValuesProblems} middle).
Note that approaches that explicitly rely on curvature integrals over the area enclosed by closed curves in shape space~\cite{Hatton:2015:NNL} do not allow for a treatment of $\chi$-isoholonomic boundary conditions.

%%%%%%%%%%%%%%%%%%%%%%%%%%%%%%%%%%%%%%%%%%%%%%%%%
%%%%%%%%%%%%%%%%%%%%%%%%%%%%%%%%%%%%%%%%%%%%%%%%%
\section{Numerical experiments}
\label{sec:Results}
We examine our model by considering optimal motion trajectories across a range of scenarios, 
including biological systems such as snakes and spermatozoa, as well as low-dimensional benchmarks like Purcell's swimmer. 
To this end, we begin with a brief discussion of the numerical discretization.
For the full details see \cref{appsec:Discretization} and \cref{appsec:Implementation}. Our code is available at \url{https://github.com/flrneha/OptimalGeometricLocomotion}.

\subsection{Numerical discretization}
For the discretization of inner and outer dissipation we employ principles of discrete geometry
and adapted variational integrators~\cite{Marsden:1999:IMS} for stable time integration of the motion trajectories.
This requires a suitable discretization both in space and time.
This discretization does not presuppose a particular parametrization of the space of admissible shapes.
To model outer dissipation in the discrete setting we adapt the framework of \cite{Gro23MfSC}, 
while for the discretization of inner dissipation we rely on a scheme similar to~\cite{bergou2008discrete}. 

A positioned shape at time $t$ represented by a curve $\gamma_t \in \cM$ is realized in the discrete setting by a polygonal curve with $N$ vertices 
$\gamma^1_t,\; \gamma^2_t,\ldots,\;\gamma^N_t$ connected by space edges. Tangent directions are given by 
$T_t^i = \tfrac{\gamma_t^{i+1}-\gamma_t^{i}}{\vert \gamma_t^{i+1}-\gamma_t^{i}\vert}$. They are used to define 
the discrete counterpart of the metric tensor  $\cB_{\boldsymbol{\gam}}$ (\eqref{eq:Bgamma}) on space edges from vertex $i$ to $i+1$ of the polygonal curve
and averaged on time intervals from time $t$ to $t+\tau$ for time step size $\tau$.

Taking finite differences in time $\tau^{-1}(\gamma_{t+\tau}^{i}-\gamma_{t}^{i})$  on time intervals and averaging these on space edges 
we obtain discrete velocities needed in the discretization of the outer dissipation.

We do not directly discretize the inner dissipation defined in \eqref{eq:InnerDissipation}. Instead we discretize a sum of the elastic energy \eqref{eq:InnerDissipationW} evaluated on incremental displacements $\delta\gamma_t = \gamma_{t+\tau}-\gamma_{t}$.

We first need a discrete notion of curvature, which measures the angle 
$\Theta^{i}_t=\sphericalangle(T_t^{i+1},T_t^{i})$ between successive tangent vectors. 
Taking finite differences in time, $\tau^{-1}(\Theta_{t+\tau}^{i}-\Theta_{t}^{i})$,
we obtain a discrete counterpart of $\delta\kappa$ in $\cW_{\gam}^{\text{bend}}$ (\eqref{eq:WbenWmem}).
Secondly, to discretize the strain energy we replace 
$\left\vert\partial_s \gam\right\vert$ by $N\vert \gam_t^{i+1} - \gam_t^{i}\vert$
and $\delta\left\vert\partial_s \gam\right\vert$ 
by $\tfrac{N}{\tau}\left(\vert \gam_{t+\tau}^{i+1} - \gam_{t+\tau}^{i}\vert 
- \vert \gam_t^{i+1} - \gam_t^{i}\vert \right)$ in $\cWmem$.

The so obtained discrete analogue of the energy gives rise to explicit expressions for the 
horizontal constraints (\eqref{eq:vertical-orthogonality}) in the form of \emph{discrete Euler--Lagrange equations}.

Computational efficiency is a challenge for highly resolved positioned shapes and the nonlinear constraint. 
In the supplementary code, we provide implementations of both the first- and second-order derivatives of all functions involved in the optimization. 
These enable a Newton type approach. Since the resulting optimal motion trajectories depend on the initialization, they constitute only local minima. 

For further details on the discretization and optimization procedure, we refer to \cref{appsec:Implementation}.

\begin{figure}%[b]
	\centering
	\includegraphics[width = 0.8\linewidth]{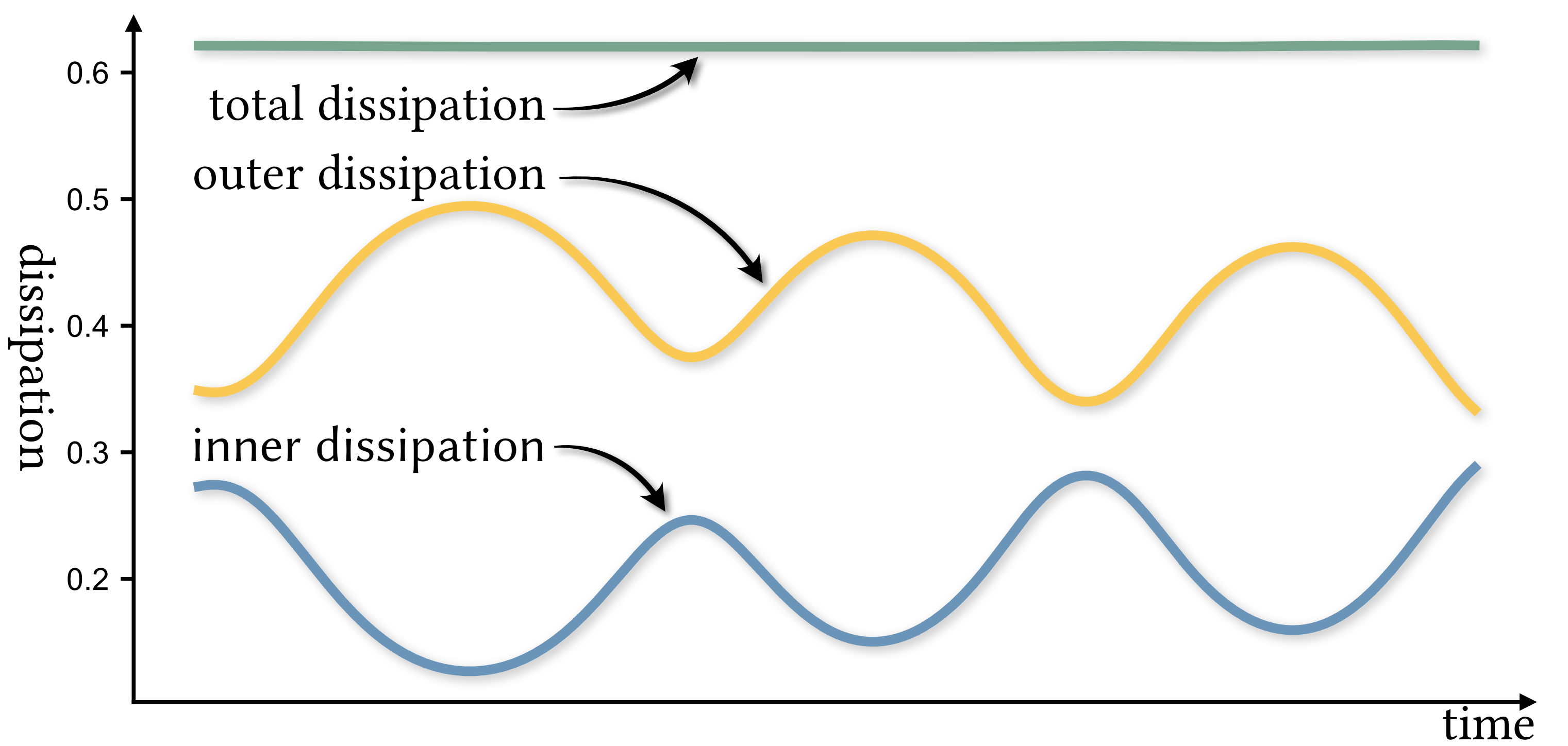}
	\caption{Graphs of the inner (blue), outer (yellow) and total (green) dissipations in time 
		for the solution of a sub-Riemannian geodesic boundary value problem.}
	\label{fig:ConstantDissipationGraph}
\end{figure}

Our experiments show quadratic convergence of the discrete solution to the continuous solution in the limit of both spatial and temporal refinement.
This renders our model capable of identifying optimal motion paths for highly resolved input shapes.
High spatial and temporal resolution is in particular necessary to properly reflect fundamental physical properties such as constant total dissipation 
along sub-Riemannian geodesic paths (\cf~\figref{fig:ConstantDissipationGraph}).
A detailed description of our optimization procedure can be found in \cref{appsec:Implementation}.
%%%%%%%
%%%%%%%

%%%%%%%%%%%%%%%%%%%%%%%%%%%%%%%%%%%%%%%%%%%%%%%%%
%%%%%%%%%%%%%%%%%%%%%%%%%%%%%%%%%%%%%%%%%%%%%%%%%
\subsection{Comparison to reduced-order models}
\begin{figure}%[b]
	\centering
	\includegraphics[width = 0.8\linewidth]{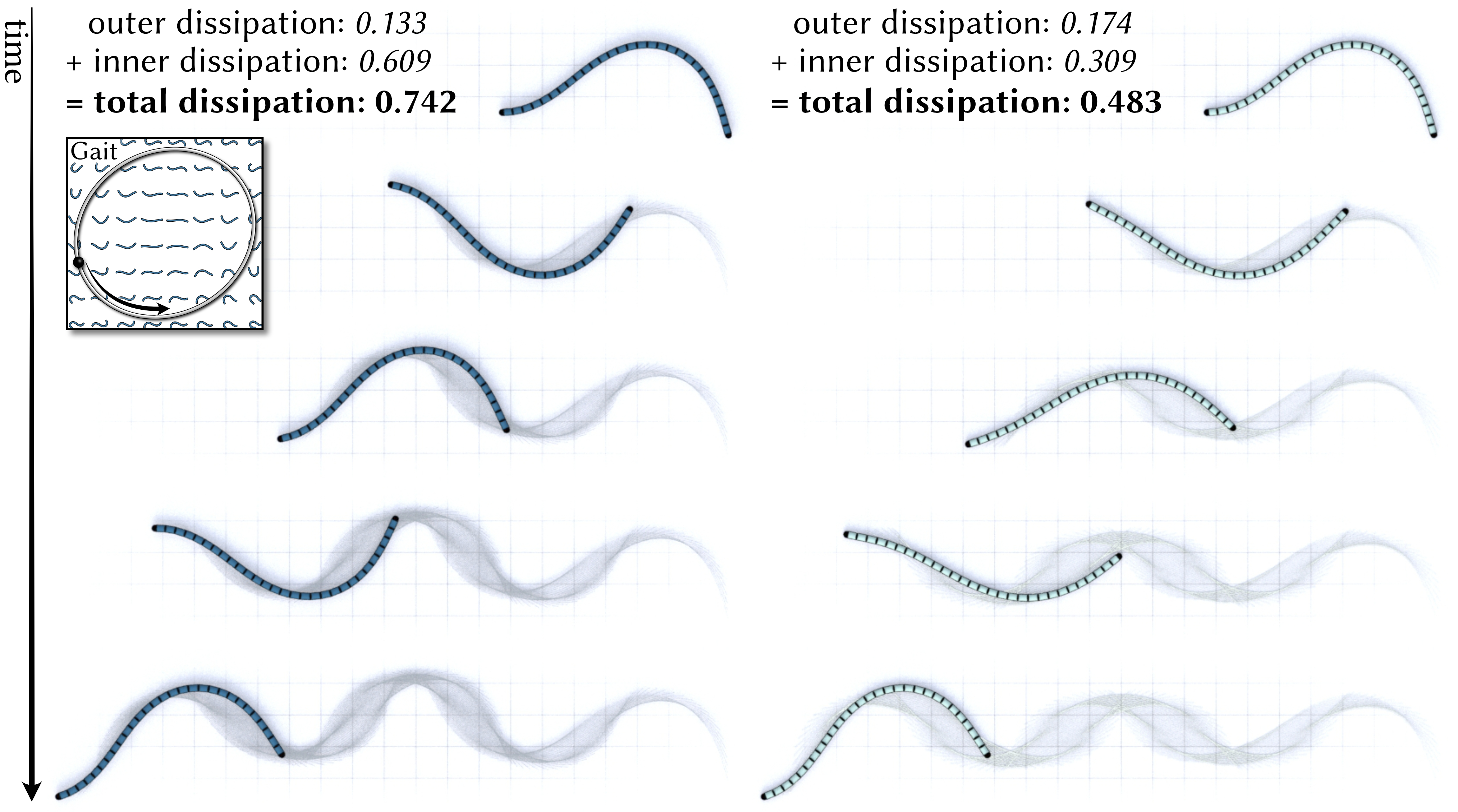}
	\caption{Left: motion path resulting form the optimal ellipse in the serpenoid shape space, right: optimal motion path as the minimizer in \eqref{eq:geodesic}
		for  initial and target body coinciding with the serpenoid boundary data. For both paths we compare the dissipation, which is significantly reduced for the minimizer of our model.}
	\label{fig:SerpenoidGaitComparison}
\end{figure}

Scalability to highly resolved shape-changing swimmers remains the prevailing computational challenge for gait optimization~\cite{Yang:2024:TGM}. 
Reduced-order models, \ie, specialized shape spaces parameterized by only a few degrees of freedom, provide a practical remedy. 
A popular example is the \emph{serpenoid shape space}, which has seen broad application owing to its good agreement with many observed locomotive 
modes
~\cite{Rieser:2024:GPp}. 
However, in our model more highly resolved discrete
shapes show lower dissipation.
In \figref{fig:SerpenoidGaitComparison} we compare two gaits.
First, we consider a motion path resulting from optimizing an elliptical gait and wavelength in the serpenoid shape space that realizes a prescribed displacement. This path is regularized by inner and outer dissipation and is obtained using the method presented in in~\cite{Becker:2025:IGL}.
Secondly, we compute the solution to the geodesic boundary value problem (\eqref{eq:geodesicPeriodic}) for initial and target body coinciding with the serpenoid boundary data. 
Notably, allowing for the high dimensional shape space the optimization leads to significantly reduced total dissipation, but also equalizes the gait's rate of 
dissipation \cref{supp-sec:Purcell}.
%%%%%%%
%%%%%%%

%%%%%%%%%%%%%%%%%%%%%%%%%%%%%%%%%%%%%%%%%%%%%%%%%
%%%%%%%%%%%%%%%%%%%%%%%%%%%%%%%%%%%%%%%%%%%%%%%%%
\subsection{Kinematic locomotors and Purcell's three link swimmer}
While the primary novelty of our work lies in its ability to handle high-dimensional systems, its evaluation on well-established low-dimen\-sional, parameterized systems remains valuable. 
Arguably, \emph{Purcell's swimmer} stands as the most prominent example~\cite{Purcell:1977:LLR}. With two degrees of freedom, it represents the simplest
model of a locomotor capable of achieving net displacement in
dissipation-dominated environments and the study of its optimal gaits has received considerable attention in the literature~\cite{Becker:2003:SPM, Tam:2007:OSP, Giraldi:2015:ODP, Kadam:2016:CPS, Hatton:2017:KCE, Alouges:2019:EOS}. 

\begin{figure}%[h]
	\centering
	\includegraphics[width = \linewidth]{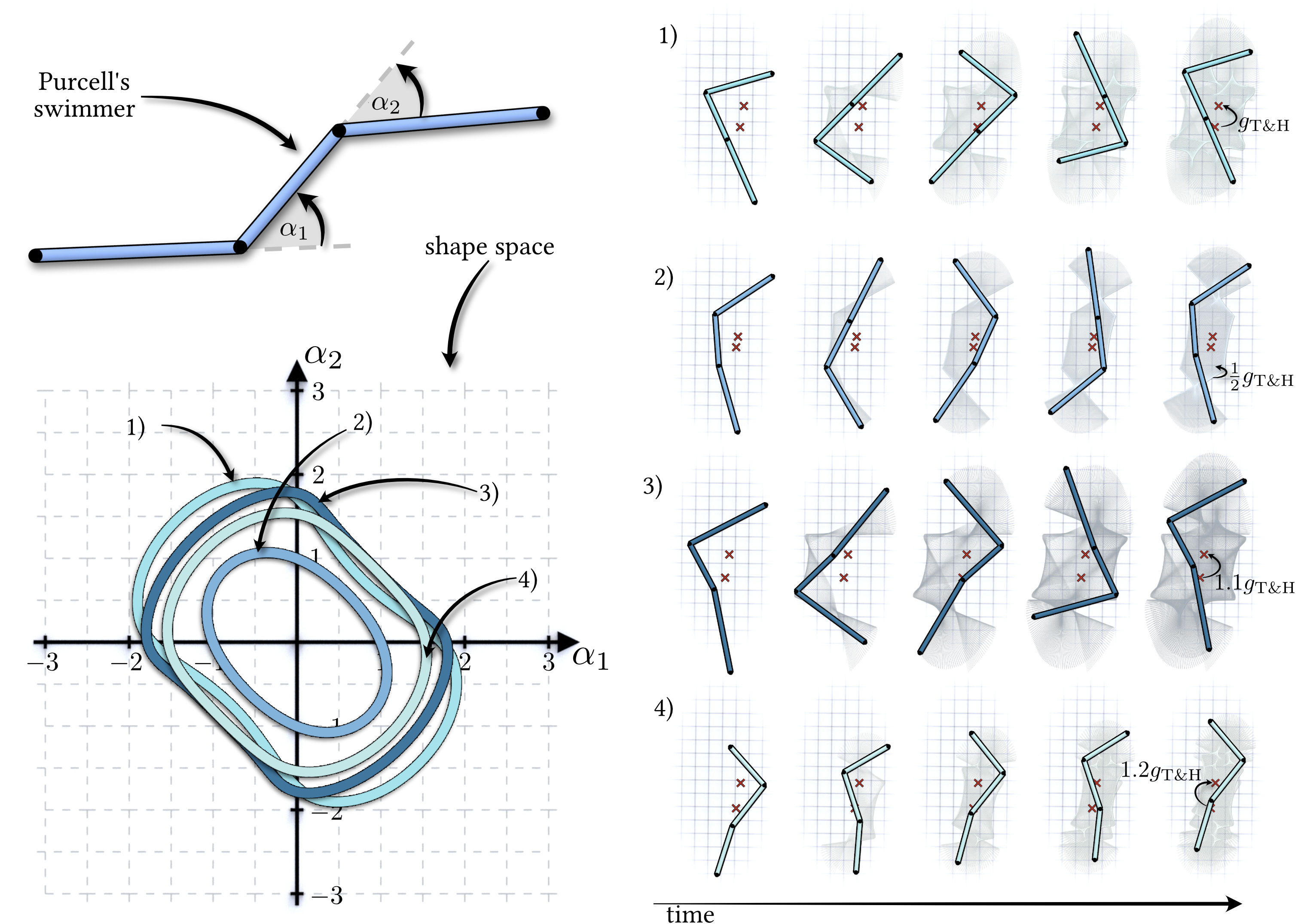}
	\caption{\figloc{Top left:} Purcell's swimmer is comprised of three consecutive edge segments parameterized by two angles and possibly edge lengths. \figloc{Bottom left:} A collection of gaits of Purcell-like swimmers in the \((\alpha_1,\alpha_2)\) shape space, representing their joint angles.  
		\figloc{Right:} The corresponding motion trajectories obtained as solutions to the isoholonomic gait problem with periodic boundary conditions. \figloc{1)} The maximal displacement gait identified in \cite{Tam:2007:OSP}, recovered here as a sub-Riemannian geodesic. \figloc{2)} Optimal gaits for smaller displacements exhibit distinct characteristics when inner energy expenditures are taken into account. \figloc{3)} Swimmers with variable arm lengths can achieve larger displacements than the classical example. \figloc{4)} For large displacements, our method identifies both the optimal gait and the optimal number of cycles required to reach the desired displacement.}
	\label{fig:Purcell}
\end{figure}

Variants of this swimmer have been
parameterized by, for instance, specifying the lengths of its ``arms''. 
For the qualitative evaluation of optimal gaits for Purcell's swimmer
in our model, we solve the isoholonomic gait problem with periodic boundary conditions. 
\figref{fig:Purcell} depicts a selection, including the maximal displacement gait identified in \cite{Tam:2007:OSP}, which we reproduce for the optimal displacement \(\dTH\), while disregarding inner dissipation and fixing the edge lengths according to the original experiments therein.
%%%%%%%
%%%%%%%

We have computed optimal gaits for a variety of target displacements \(g\), as presented in \cref{supp-sec:Purcell}.
In addition we allow for a variation of the length of the swimmer's limbs regularized by the inner dissipation.
In close agreement with the findings in \cite{Tam:2007:OSP}, the optimal shape sequence exhibits characteristic non-convex indentations when visualized in the shape space parameterized by the joint angles \((\alpha_1,\alpha_2)\). 

If we prescribe a displacement of only \(\tfrac{1}{2}\dTH\), we find that inner dissipation has a significant influence on the optimal gaits.
When efficiency measures are incorporated, optimal gaits form convex, ellipse-like shapes in the shape space consistent with previous findings ~\cite{Tam:2007:OSP}.
Solutions to \teqref{eq:geodesicPeriodic} implicitly encode the optimal number of periodic cycles in shape space required to achieve the target displacement \(g\).
Our model automatically chooses an optimal number of %periodic 
cycles to reach a desired displacement. 
Note that, even when solutions to \teqref{eq:geodesicPeriodic} traverse a fiber multiple times, 
they are not necessarily the horizontal lift of a multiply traversed closed loop in shape space.
Remarkably, we achieve displacements larger than \(\dTH\) for generalized Purcell's swimmer with variable limb length.

%%%%%%%%%%%%%%%%%%%%%%%%%%%%%%%%%%%%%%%%%%%%%%%%%
%%%%%%%%%%%%%%%%%%%%%%%%%%%%%%%%%%%%%%%%%%%%%%%%%
\subsection{Undulating geometric locomotion across length scales}
Observed across all length scales, geometric locomotors in nature frequently move by means of body undulations~\cite{Rieser:2024:GPp}.
%%%%%%%
\begin{figure}%[h]
	\centering
	\includegraphics[width = 0.8\linewidth]{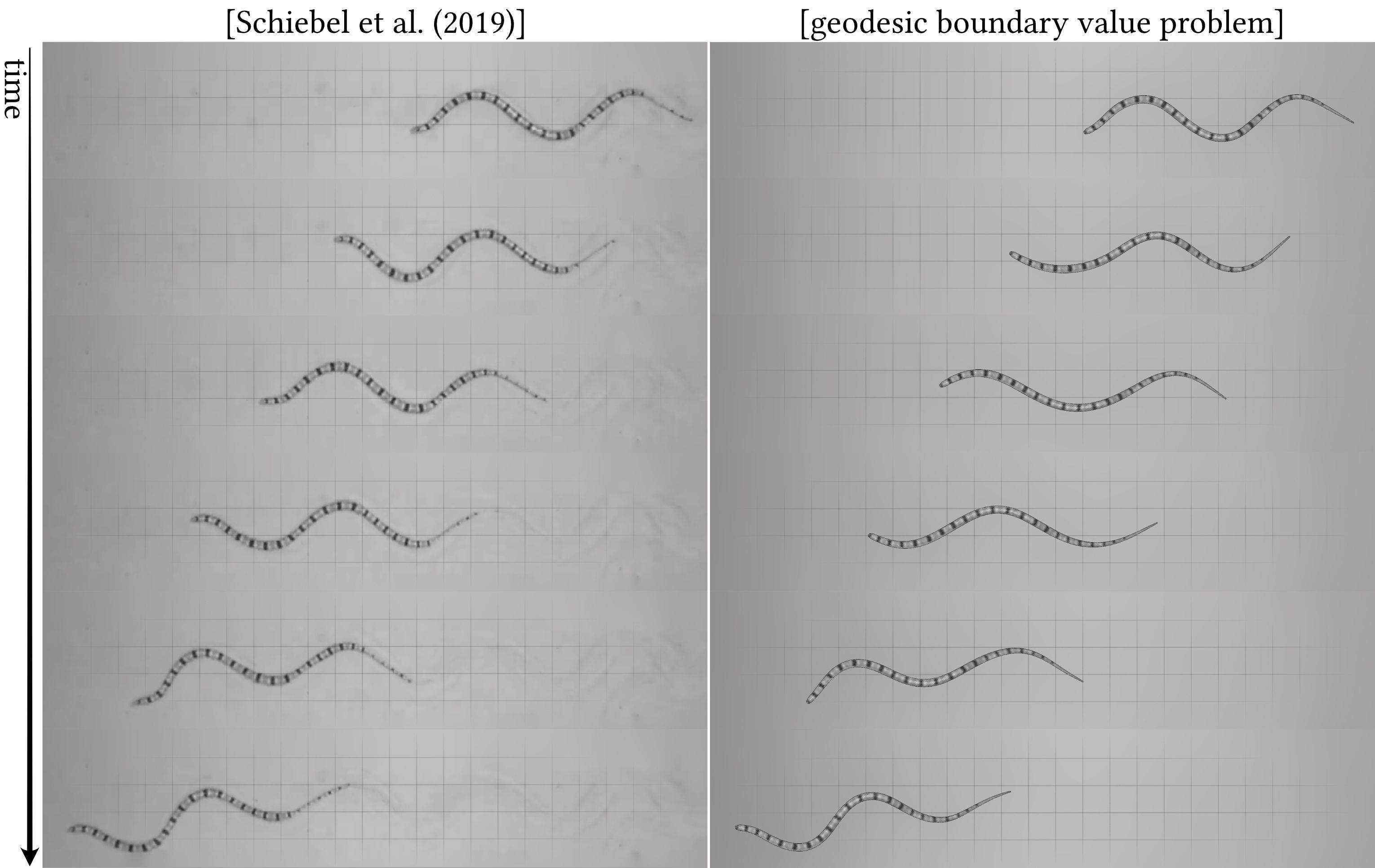}
	\caption{\figloc{Left:} The motion of a shovel nosed snake adapted from the supplementary materials provided in \cite{Schiebel:2019:MDR}. \figloc{Right:} Solution to the sub-Riemannian geodesic boundary value problem with boundary configurations matching the video data.}
	\label{fig:GeorgiaComparison}
\end{figure}
%%%%%%%
One example is given by the  shovelnose snake \emph{(Chionactis occipitalis)}, slithering on or buried within highly dampening granular media, such as sand. 
In \figref{fig:GeorgiaComparison} we show the sub-Riemannian geodesic computed for fixed initial and target body, which are extracted from data provided in~\cite{Schiebel:2019:MDR}. We observe a good qualitative agreement with the original data. In \cite[Figure 2]{Rieser:2024:GPp}, the authors show that the snake prefers larger amplitudes and higher curvatures when buried, compared to moving on the surface only. 
The increase of curvature under stronger outer dissipation, as one would expect for buried snake, is well-predicted by our model (\figref{fig:WeightsComparison}).

\begin{figure}%[b]
	\centering
	\includegraphics[width = .9\linewidth]{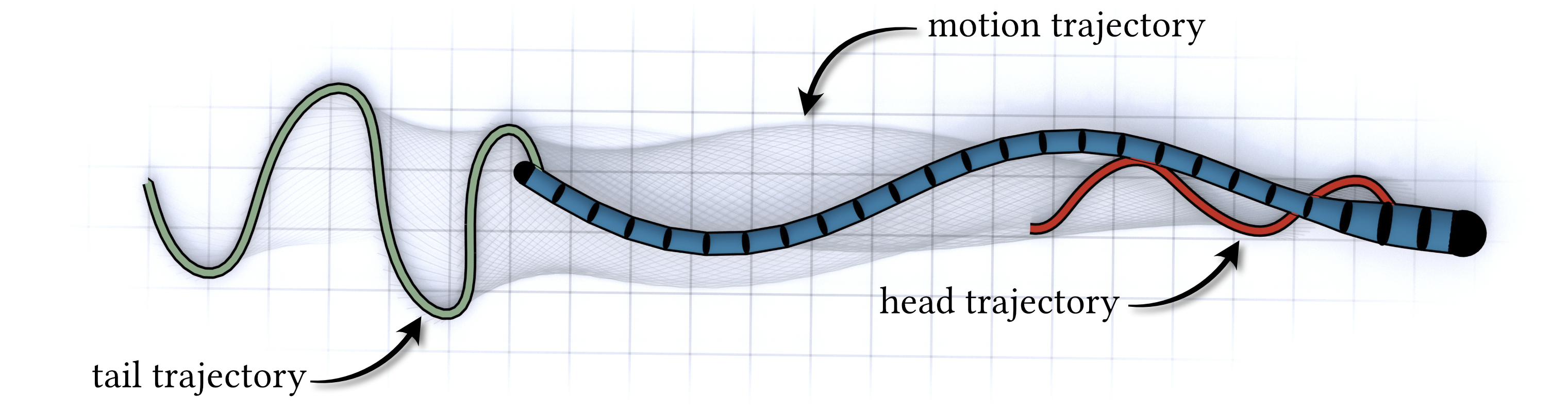}
	\caption{Optimal motion trajectory of a slender locomotor with non-uniform thickness and proportionally varying outer dissipation weight. 
		Notably, the motion shows different amplitudes in the trajectories of the head \figloc{(red)} and tail \figloc{(green)} segments.}
	\label{fig:SpermCell}
\end{figure}

Other slender locomotors, such as spermatozoa, exhibit a pronounced non-uniformity of body thickness.
Such effects can be readily incorporated into our model by increasing the outer dissipation weight in thicker regions. In \figref{fig:SpermCell}, we demonstrate that 
computed optimal motion paths for the isoholonomic gait problem (\teqref{eq:geodesicPeriodic}) exhibit different amplitudes in the trajectories of the head and tail segments. This outcome is in good qualitative agreement with observations from natural organisms~\cite{Guasto:2020:FKR, Yaqoob:2023:MOU}.
%%%%%%%
%%%%%%%

\begin{figure}[h]
	\centering
	\includegraphics[width = 0.8\linewidth]{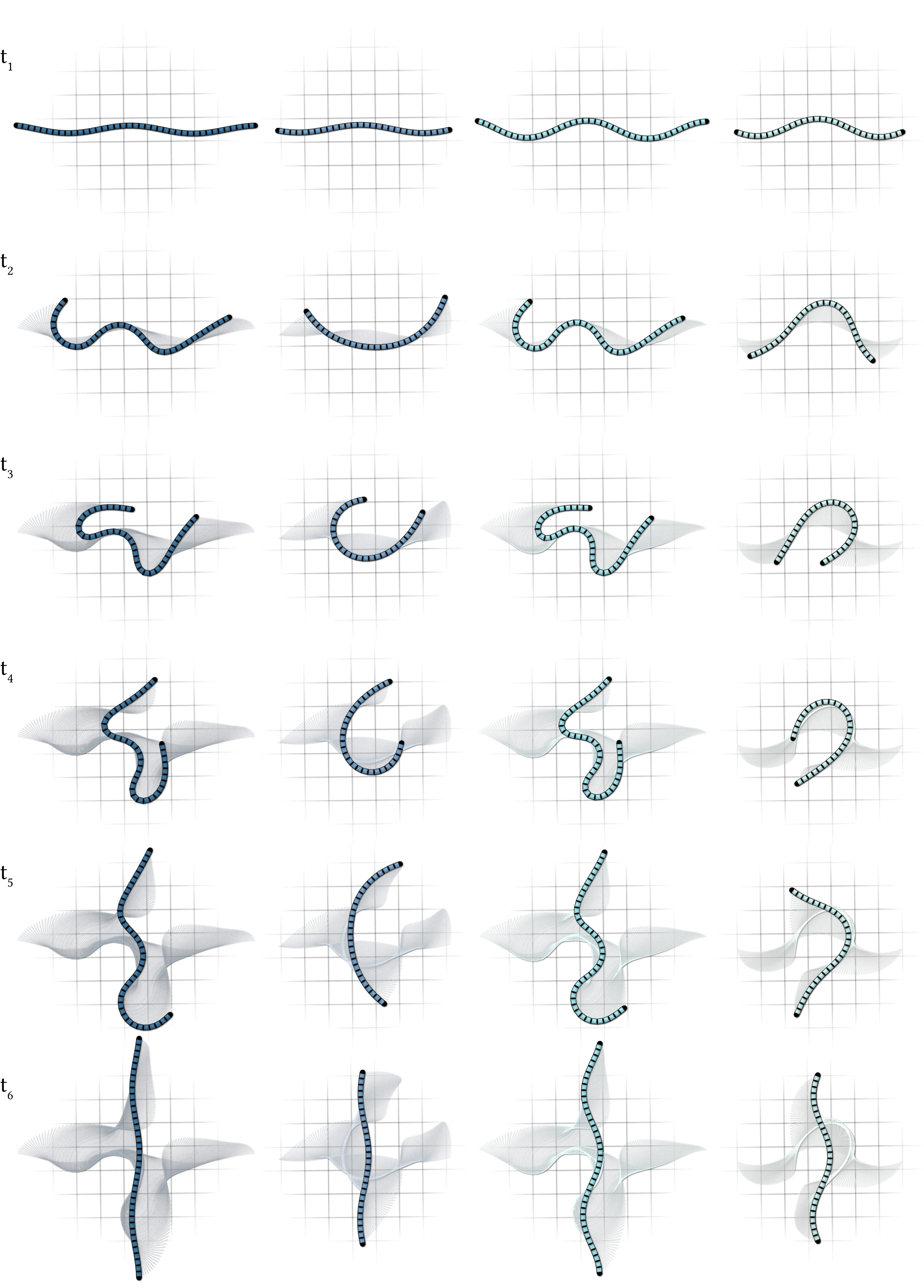}
	\caption{Different optimal turning strategies for a  \(90^\circ\)-rotation 
		with initial shape of different length and curviness for six equidistant time steps $t_1, \ldots, t_6$.}
	\label{fig:TurningGait}
\end{figure}

The need to find optimal motion trajectories extends beyond mere
forward locomotion. Turning gaits, in particular, remain an active
area of research in fields such as biomechanics and robotics
alike~\cite{Kim:2011:TST, Kohannim:2012:OTG, Wang:2020:TOT,
	Rozemuller:2024:SOC}. In \figref{fig:TurningGait}, we depict optimal
turning gaits for a slender locomotor based on the boundary value
problem \eqref{eq:ChiHolonomicProblem}. Our experiments identify
turning strategies that closely resemble (variants of)
\(\Omega\)-turns as optimal for \(90^\circ\)-rotations---regardless of
the length or curvature of the snake.

\section{Conclusion}
The present work introduces a geometric optimization approach for shape change driven locomotion of slender bodies.
It is distinguished by accounting for both inner and outer dissipation. 
We optimize total dissipation subject to the horizontality constraint 
in the frame work of sub-Riemannian geometry
for three different types of application relevant 
boundary conditions.

Our model considers bodies as elements in the infinite dimensional space of curves
and approximates them with polygonal discretizations of variable dimensionality.
It reproduces the qualitative motion of slender planar locomotors observed in nature such as snakes and spermatozoa. 
As a special case in the low dimensional extreme our model recovers known optimal strategies for Purcell's swimmer, extending and improving upon them in more general settings. 

A  limitation of our model in its present form is that it does not explicitly prevent self-intersections of the 
shape changing body, which limits its immediate applicability to robotic systems. 
Therefore, a promising direction for future work would be to couple our geometric locomotion model with 
collision avoidance mechanisms on shape space, as explored in, \eg,~\cite{yu2021repulsive}.

Generalizing our computational framework for an outer dissipation, which depends nonlinearly on the strain, would allow us to address 
systems such as \(N\)-sphere swimmers~\cite{Moreau:2023:COC}.
Another possible, theoretically straightforward, extension is the adaptation of our  framework to surface- or skeleton-based shape representations.
Moreover, in inertia-dominated scenarios, augmenting the kinetic energy metric on configuration space with our internal dissipation term allows for studying optimal reorientation strategies for astronauts in microgravity or classic examples such as the ``falling cat'' problem.

\section*{Acknowledgments}
This work was supported by the Deutsche Forschungsgemeinschaft
(DFG, German Research Foundation) via project project 539309657 -- Collaborative Research Center 1720, and via Germany’s Excellence Strategy
project 390685813 -- Hausdorff Center for Mathematics. Additional support was provided through Houdini software, courtesy of SideFX.

\printbibliography
\appendix
\markboth{Appendix}{Appendix}

\section{Derivation of the horizontality constraint} \label{supp:horizconstraint}
In this section, we derive the horizontality constraint for a motion path in a general setting.
For a Riemannian manifold $(\mathcal{M},\langle\cdot,\cdot\rangle)$ 
let $\boldsymbol{\gam}\colon[0,1]\to \mathcal{M}$ be a smooth path with finite energy
$
\mathcal E(\boldsymbol{\gam})
= \frac12\int_0^1 \langle\boldsymbol{\gam}_t',\boldsymbol{\gam}_t'\rangle \, dt
$
and let  $\Gamma\colon(-\varepsilon,\varepsilon)\times[0,1]\to \mathcal{M}$ be a smooth
variation of $\boldsymbol{\gam}$ with $\Gamma(0,t)=\boldsymbol{\gam}_t$ and the infinitesimal variation 
$\delta\boldsymbol{\gam}_t=\partial_s \Gamma(0,t)$.
We have that the Lie bracket of coordinate vector fields vanishes, \ie, $[\partial_t\Gamma,\partial_s\Gamma]=0$, which restricted to $s=0$ gives $[\boldsymbol{\gam}',\delta\boldsymbol{\gam}]=0$
along $\boldsymbol{\gam}$.

The first variation of the energy in the direction of the  variation $\delta\boldsymbol{\gam}$ of the path $\boldsymbol{\gam}$ is
\[
d\mathcal E(\boldsymbol{\gam})(\delta\boldsymbol{\gamma})
= \frac12\int_0^1 d\langle\boldsymbol{\gam}_t',\boldsymbol{\gam}_t'\rangle(\delta\boldsymbol{\gamma}_t)\, dt
= \int_0^1 \langle\nabla_{\delta\boldsymbol{\gamma}_t}\boldsymbol{\gam}_t',\boldsymbol{\gam}_t'\rangle \, dt ,
\]
where we use that the Levi--Civita connection preserves the metric.
Since the Levi--Civita connection is also torsion free, we have
$
\nabla_{\delta\boldsymbol{{\gamma}}}\boldsymbol{\gam}' - \nabla_{\boldsymbol{{\gamma}'}}\delta\boldsymbol{\gamma}
= [\delta\boldsymbol{\gamma},\boldsymbol{\gamma}']=0,
$
and thus $\nabla_{\delta\boldsymbol{\gamma}}\boldsymbol{\gamma}'   = \nabla_{\boldsymbol{\gamma}'}\delta\boldsymbol{\gamma}$.
Taking into account that
$\tfrac{\mathrm{d}}{\mathrm{d}t} \langle\delta\boldsymbol{\gamma}_t,\boldsymbol{\gamma}_t'\rangle =
\langle\nabla_{\boldsymbol{\gamma}_t'}\delta\boldsymbol{\gamma}_t,\boldsymbol{\gamma}_t'\rangle + \langle\delta\boldsymbol{\gamma}_t,\nabla_{\boldsymbol{\gamma}_t'}\boldsymbol{\gamma}_t'\rangle
$  
we obtain
\begin{equation}
	d\mathcal E(\boldsymbol{\gamma}_t)\delta\boldsymbol{\gamma}_t
	= \langle\delta\boldsymbol{\gamma}_t,\boldsymbol{\gamma}_t'\rangle\big|_{t=0}^{t=1}
	- \int_0^1 \langle\delta\boldsymbol{\gamma}_t,\nabla_{\boldsymbol{\gamma}_t'}\boldsymbol{\gamma}_t'\rangle \, dt ,
	\label{eq:FirstVariation}
\end{equation}
%%%%%%%%%%%%%%%%%
Next, we consider the setting of the main text.  We have a principal fiber bundle $\pi\colon\mathcal M\to\mathcal S$ underlying the Riemannian
manifold $(\mathcal M,\langle\cdot,\cdot\rangle)$ with structure group $G$.
The group $G$ acts on $\mathcal M$ by rigid body transformations, and the
metric $\langle\cdot,\cdot\rangle$ is $G$ invariant. 
We refer to the infinitesimal generators of the action of $G$ on $\mathcal M$ 
as vertical vector fields and denote the space of vertical vector fields along 
$\boldsymbol{\gam}$ by $V_{\boldsymbol{\gam}}$. 
For a stationary point $\boldsymbol{\gam}$ of the energy under vertical variations $\delta\boldsymbol{\gamma}$
the invariance of the metric implies that $\delta\boldsymbol{\gamma}_t$ is a Killing vector field on $\mathcal M$.

Now, we have to show  
that a path $\boldsymbol{\gam}$ in $\mathcal M$ is a stationary point of the energy
under temporally smooth vertical variations $\delta\boldsymbol{\gamma}\in V_{\boldsymbol{\gam}}$
if and only if, for all smooth variations
$\delta\boldsymbol{\gamma}\in V_{\boldsymbol{\gam}}$ the horizontality constraint as a 
Riemannian-Noether condition
\begin{equation*}
	\langle\boldsymbol{\gam}_t',\delta\boldsymbol{\gamma}_t\rangle = 0 
\end{equation*}
holds for all $t \in [0,1]$. 

To see this, we first assume that $\boldsymbol{\gam}$ in $\mathcal M$ is a stationary point of the energy
under temporally smooth vertical variations $\delta\boldsymbol{\gamma}\in V_{\boldsymbol{\gam}}$. Restricting to variations 
that vanish for $t=0$ and $t=1$ we obtain from \teqref{eq:FirstVariation} 
by the fundamental lemma that 
\(\nabla_{\boldsymbol{\gam}_t'}\boldsymbol{\gam}_t' \perp_\cM  V_{\boldsymbol{\gam}_t}\) for all $t\in (0,1)$. 
Now, considering $\delta\boldsymbol{\gamma}$  that vanish at \(t=1\) we in addition find \(\boldsymbol\gam'_0\perp_\cM V_{\gam_0}\).
The invariance property of the metric w.r.t. rigid body motions implies that the variation $\delta\boldsymbol{\gamma}_t$
is a Killing vector field for all $t\in [0,1]$ and thus in particular 
$\langle\nabla_{\boldsymbol{\gamma}_t'}\delta\boldsymbol{\gamma_t},\boldsymbol{\gamma}_t'\rangle = 0$
for all $t\in [0,1]$.
This implies 
\begin{equation}
	\tfrac{\mathrm{d}}{\mathrm{d}t} \langle\delta\boldsymbol{\gamma}_t,\boldsymbol{\gamma}_t'\rangle
	= \langle\nabla_{\boldsymbol{\gamma}_t'}\delta\boldsymbol{\gamma}_t,\boldsymbol{\gamma}_t'\rangle
	+ \langle\delta\boldsymbol{\gamma}_t,\nabla_{\boldsymbol{\gamma}_t'}\boldsymbol{\gamma}_t'\rangle =0
	\label{eq:KillingRelation}
\end{equation}
and using \(\boldsymbol\gam'_0\perp_\cM V_{\gam_0}\) the claim 
$\langle\boldsymbol{\gam}_t',\delta\boldsymbol{\gamma}_t\rangle = 0$ 
follows for all $t\in [0,1]$.

To show the reverse implication, assume that $\langle\boldsymbol{\gam}_t',\delta\boldsymbol{\gamma}_t\rangle = 0$ 
for all $t\in [0,1]$. Then, \eqref{eq:KillingRelation}  together with the Killing property of $\delta\boldsymbol{\gamma}$ implies 
$
0=  \langle\delta\boldsymbol{\gamma}_t,\nabla_{\boldsymbol{\gamma}_t'}\boldsymbol{\gamma}_t'\rangle
$
for all $t\in (0,1)$.
From this and $\langle\boldsymbol{\gam}_t',\delta\boldsymbol{\gamma}_t\rangle = 0$ for $t\in\{0,1\}$ we obtain
$ d\mathcal E(\boldsymbol{\gamma})\delta\boldsymbol{\gamma}=0$ for all 
$\delta\boldsymbol{\gamma}\in V_{\boldsymbol{\gamma}}$.

%%%%%%%%%%%%%%%%%%%%%%%%%%%%%%%
%%%%%%%%%%%%%%%%%%%%%%%%%%%%%%%
%%%%%%%%%%%%%%%%%%%%%%%%%%%%%%%
\section{Discretization}\label{appsec:Discretization}
For our numerical experiments, we discretize our model in both space and time. 
For the discretization of inner and outer dissipation we employ principles of discrete geometry and adapted variational integrators~\cite{Marsden:1999:IMS} for stable time integration of the motion trajectories.

We consider discrete slender body shapes \( \gamma \in \sfM\) in the discrete configuration space \(\sfM\) of polygonal curves with \(N\) vertices \({\gam^1},\ldots,{\gam^N} \). We can consider a vectorized representation
\begin{equation*}
	\gam = {({\gam^1},\ldots,{\gam^N})}^\top\in\RR^{3N}.
\end{equation*}
Each configuration \(\gam\in \RR^{3N}\) is specified by the shape's vertex positions embedded in \(\RR^3\). The shape space is defined as the quotient \(\sfS = \sfM/G\) and does not require an explicit parameterization in our sub-Riemannian framework.
We consider motion paths with discrete timepoints \(t_k = k \tau\) with \(k = 0, \ldots, K\) and \(\tau = \tfrac{1}{K}\).
Consequently, a time-discrete motion is described by the ordered sequence of configurations \(\boldsymbol\gamma = (\gamma_{t_0},\ldots,\gamma_{t_K})\).

For each edge of a shape \(\gamtk\) we define a unit tangent vector as
\begin{equation*}
	T_\tk^{i} \coloneqq \tfrac{\gamma_\tk^{i+1}-\gamma_\tk^i}{|\gamma_\tk^{i+1}-\gamma_\tk^i|}.
\end{equation*}
Moreover, we denote the \(N-1\) edge midpoints and their corresponding velocities at time \(t_k\), respectively, by
\begin{equation*}
	\eta_{t_k}^{i} = \tfrac{1}{2}(\gam_\tk^{i+1}+\gam_\tk^i) \qquad \text{and} \qquad \nu_{t_k}^i = \tfrac{1}{\tau} (\eta_{t_{k+1}}^i-\eta_{t_k}^i) \, .
\end{equation*}

\subsection{Discrete total dissipation}
\label{suppsec:DiscreteLagrangian}
In the continuous setting, the energy Eq.\,(1) measures total energy dissipation as the energy of a motion path \(\boldsymbol \gamma\) in the configuration space \(\sfM\). Mimicking this definition, the \emph{discrete energy}
\begin{equation} \label{eq:app:dscreteEnergy}
	\sfE(\boldsymbol{\gamma}) = \tfrac{1}{2} ( \sfE_\dext(\boldsymbol\gam) + \sfE_\dint(\boldsymbol\gamma))
\end{equation}
is defined as the sum of discrete inner and outer dissipation energies measured by the discrete dissipations induced by the discrete displacements of \(\boldsymbol\gamma\) between consecutive time steps of duration \(\tau = \tfrac{1}{K}\). As in the continuous formulation, we model this dissipation as quadratic forms associated to discrete analogues of the inner and outer dissipation metrics (Eqs.\,(6) and~(9), respectively). To model outer dissipation in the discrete setting we adapt the framework proposed in~\cite{Gro23MfSC} to accommodate edge-based dissipation, while for the discretization of inner dissipation we follow~\cite{Heeren:2012:TDG, Heeren+:2014:EtG} adapted to polygonal curve geometries.

\subsubsection*{Discrete outer dissipation}
As explained in the main text, in practice, one can approximate outer dissipation using \emph{resistive force theory}~\cite{Gray:1955:PSU,Zhang:2014:TEO}. Ignoring long-range interactions, this theory models total energy loss by summing the dissipation from individual dissipation elements. We use a discrete version of~Eq.\,(7).

The displacement of an edge in a viscous medium dissipates energy to the environment, which we capture via local dissipation tensors
\begin{equation*}
	\sfB_\tk^{i} = \vert \gamma_\tk^{i+1} - \gam_\tk^i \vert (I + (\epsilon_{i} -1) (T_\tk \otimes T_\tk)^{i}) \in \RR^{3 \times 3}.
\end{equation*}
Here, the \emph{local anisotropy ratio}
\(\epsilon_{i}\in(0,1]\) controls directional friction: for \(\epsilon_{i}\approx 0\), tangential motion incurs negligible dissipation, while for \(\epsilon_{i} =1\), the tensors become isotropic. 

Assembling these tensors for all individual edges into block--diagonal matrices \(\sfB_{t_k}\in\mathbb{R}^{3(N-1) \times 3(N-1)}\), we define a dissipation tensor that is averaged on time edges
\begin{equation*}
	\bar{\sfB}_\tk=\tfrac12(\sfB_\tk+\sfB_\tkk)
\end{equation*}
to discretize the maps in Eq.\,(5).
This yields a \(G\)--invariant, symmetric, positive-definite \emph{discrete outer dissipation metric} on \(\mathbb{R}^{3(N-1)}\) and the outer dissipation contributes to the discrete Lagrangian with the  quadratic form
\begin{align}
	\label{eq:DiscreteOuterDissipation}
	\sfE_\dext(\boldsymbol\gamma) &=\tau\sum_{k=0}^{K-1}\langle\bar{\sfB}_\tk \nu_\tk,\nu_\tk\rangle_{\mathbb{R}^{3(N-1)}} \nonumber \\
	& = K\sum_{k=0}^{K-1}\langle\bar{\sfB}_\tk \eta_\tkk -\eta_\tk,\eta_\tkk - \eta_\tk \rangle_{\mathbb{R}^{3(N-1)}} \,.
\end{align}
When the context allows for it, we will omit the index indicating the metric in \(\RR^{3(N-1)}\) for simplicity.
\subsubsection*{Discrete inner dissipation}
In the continuous setup, instantaneous strain densities give rise to a Riemannian structure on the space of thin elastic rods. 
Following established discretization schemes~\cite{Heeren:2012:TDG, Heeren+:2014:EtG}, we discretize the inner dissipation energy \(\cW\) (Eq.\,(8)).
To discretize the strain energy \(\cW_\mem\) we use the discrete length element
$N\vert \gam_\tk^{i+1} - \gam_\tk^{i}\vert$
and its discrete derivative 
$\tfrac{N}{\tau}\left(\vert \gam_\tkk^{i+1} - \gam_\tkk^{i}\vert 
- \vert \gam_\tk^{i+1} - \gam_\tk^{i}\vert \right)$ in $\sfW_\mem$. 
For a discrete version of \(\cW_\ben \) we consider a discrete notion of curvature measuring it by the change of angles  \(\Theta_\tk^i =\sphericalangle(T_\tk^{i+1},T_\tk^{i}) \) between two consecutive tangent vectors~\cite{Carroll:2014:ASD}.
Together this results in

\begin{align*}
	\sfW(\gam_\tk,\gam_\tkk) &= \sfW_\mem(\gam_\tk,\gam_\tkk) + \sfW_\ben(\gam_\tk,\gam_\tkk)\\ %,
	&= \tfrac{1}{\tau^2}\sum_{i=0}^{N-1}\left(1 -\tfrac{|\gamma_\tkk^{i+1}-\gamma_\tkk^i|}{|\gamma_\tk^{i+1}-\gamma_\tk^i|}\right)^2 %= 
	+ \tfrac{1}{\tau^2}\sum_{i=1}^{N-1}\tfrac{(\Theta_\tkk^i-\Theta_\tk^i)^2}{|\gamma_\tk^{i+1}-\gamma_\tk^i|}.
\end{align*}
Discretely integrating this in time results in the inner dissipation contribution to the discrete Lagrangian
\begin{equation*}
	\sfE_\dint(\boldsymbol\gamma)
	=K\sum_{k=0}^{K-1}\sfW(\gamtk,\gamtkk).
\end{equation*}
\subsection{Discrete horizontality constraints}
Our sub-Riemannian formulation requires that critical motion paths are horizontal geodesics in the configuration space. This condition is derived via the Euler-Lagrange equation for the energy under vertical variations. Notably, because the vertical subspace field coincides with the kernel of the inner dissipation metric, this condition only depends on the outer dissipation metric (Eq.\,(6)). This transfers to the discrete case, where the discrete inner dissipation energy \(\sfW\) also vanishes under vertical variations.
Hence, following the principle to \emph{first discretize, then optimize} we derive a discrete analogue of the corresponding Euler-Langrange equations considering vertical variations of the discrete outer dissipation energy \(	\sfE_\dext \) (\teqref{eq:DiscreteOuterDissipation}).

We can parameterize vertical variations of the shape \(\gamtk\) as 
\[
\delta\gam_\tk = \omega\times \gamtk + b.
\]
In the following, we use the notation \(\delta X \) for the variational derivative of any quantity \(X\) that depends on \(\gam_\tk \) in the direction \(\delta \gam_\tk\) \ie \,
\begin{equation*}
	\delta X = dX_\tk(\delta \gam_\tk) = \left. \tfrac{d}{d \epsilon}(X(\gam_\tk + \epsilon \delta \gam_\tk))\right\vert_{\epsilon=0}
\end{equation*}
Then, since \(\delta \nu_\tk = - \delta \eta_\tk\),
We find 
\begin{equation*}
	d\sfE_\dext(\boldsymbol \gam)(\delta\gam_\tk) 
	= \tfrac{1}{4} \langle \delta \sfB_\tk \nu_\tk, \nu_\tk\rangle + \langle \bar{\sfB}_\tk \nu_\tk, \delta \nu_\tk \rangle = \tfrac{1}{4} \langle \delta \sfB_\tk \nu_\tk, \nu_\tk \rangle - \langle \bar{\sfB}_\tk \nu_\tk, \delta \eta_\tk\rangle.
\end{equation*}

\paragraph{Case \(\omega = 0\):}
Sine $\sfB_\tk$ is invariant under translations, we immediately see for variations with \(\omega=0\) that \(d\sfE_\dext(\boldsymbol \gam)(\delta\gam_\tk) =0\) for all those variations if and only if 
\[
\left\langle - \sum_{i} \bar{\sfB}_\tk^{i} \nu_\tk^{i}, b\right\rangle = 0
\]
for any \(b\in\RR^3\).
\paragraph{Case \(b = 0\):}
In \cite{Gro23MfSC} it is shown that for (in block diagonal notation) the variations of \(\sfB_\tk\) it holds \(\delta \sfB_\tk = [[[\omega]_\times], \sfB_\tk]\) with $ 
[[\omega]_\times] \in \RR^{3n, 3n}$ the block diagonal matrix with all blocks equal to the cross product matrix $[\omega]_\times \in \RR^{3,3}$.

Hence, due to the symmetry of \(\sfB_\tk\) and the skew-symmetry of \([[\omega]_\times]\),
\[\tfrac{1}{2}\langle \delta \sfB_\tk \nu_\tk, \nu_\tk \rangle = - \tfrac{1}{2}\langle [[[\omega]_\times], \sfB_\tk]\nu_\tk, \nu_\tk\rangle = -\langle  \sfB_\tk \nu_\tk,[[\omega]_\times] \nu_\tk \rangle.\]
Edge-wise this is expressed as 
\[\tfrac{1}{4} \langle \delta \sfB_\tk \nu_\tk, \nu_\tk\rangle =-\tfrac{1}{2}\sum_{i} \langle \sfB_\tk^{i} \nu_\tk^{i}, \omega\times \nu_\tk^{i}\rangle.\]
Hence, 
\begin{align*}
	d\sfE_\dext(\boldsymbol \gam)(\delta\gam_\tk) 
	&= -\tfrac{1}{2} \langle \sfB_\tk \nu_\tk, [[\omega]_\times] \nu_\tk\rangle - \langle \bar{\sfB}_\tk \nu_\tk, [[\omega]_\times] \eta_\tk \rangle \\
	&= -\tfrac{1}{2} \left( \langle \sfB_\tk \nu_\tk,[[\omega]_\times]\eta_\tkk \rangle + \langle \sfB_\tk \nu_\tk, [[\omega]_\times]\eta_\tk \rangle \right.\\
	&\qquad\qquad \left. - \langle \sfB_\tk \nu_\tk,[[\omega]_\times] \eta_\tk\rangle + \langle \sfB_\tkk \nu_\tk, [[\omega]_\times]  \eta_\tk \rangle\right) \\
	&= -\tfrac{1}{2} \left( \langle \sfB_\tk \nu_\tk, [[\omega]_\times]\eta_\tkk \rangle + \langle \sfB_\tkk \nu_\tk, [[\omega]_\times]  \eta_\tk\rangle\right) \\
	&= \left\langle \omega, -\tfrac{1}{2}\sum_{i} \left(  \eta_\tkk^{i}\times \sfB_\tk^{i} \nu_\tk^{i} + \eta_\tk^{i}\times \sfB_\tkk^{i} \nu_\tk^{i}\right) \right\rangle 
\end{align*}
Collecting everything we conclude  that
a discrete motion path \(\boldsymbol{\gam}\subset\sfM\) is horizontal if and only if   	
\begin{align}
	\label{eq:MomentumCondition}
	\begin{pmatrix}
		-\tfrac{1}{2}\sum_{i} \left( \eta_{\tkk}^{i}\times \sfB_\tk^{i} \nu_\tk^{i} + \eta_\tk^{i}\times \sfB_\tkk^{i} \nu_\tk^{i}\right) \\
		- \sum_{i} \bar{\sfB}_\tk^{i} \nu_\tk^{i}
	\end{pmatrix} \equiv 0.
\end{align}
Note that this condition in form of six equality constraints is straightforward to enforce numerically and requires no explicit projection onto the vertical subspace field. 

\subsection{Additional constraints}
The boundary value problems introduced in Sec.\,2.2 of the main text require the handling of some additional constraints. Most of them are straightforward to implement.
For the geodesic boundary value problem (Eq.\,(12)) we just set equality constraints on initial  and target configuration \(\gam_{t_0} \) and \(\gam_{t_K}\).
In the case of periodic boundary conditions, a desired holonomy \(g \in G\) is imposed via the equality constraint 
\begin{equation*}
	\gam_{t_K} - g(\gam_{t_0} ) = 0
\end{equation*} 
For \(\chi\)-isoholonomic boundary conditions comparing initial and final configurations \(\gam_{t_0} \) and \(\gam_{t_K}\) respectively, requires a reference gauge \(\chi\) as described in Sec.\,2.2.3 of the main text. In practice, the realised rigid body motion can be found by a rigid registration of \(\gam_{t_0}\) to \(\gam_{t_K}\), which is given by \( h(\gam_{t_0})\), where \(h\) is the solution to
\begin{equation}\label{eq:rigidregistration}
	\argmin_{h\in G} \sum_{i} \vert \gam_{t_K}^{i+1} -\gam_{t_K}^i \vert ( h(\gam_{t_0}) - \gam_{t_K})^2.
\end{equation}
The discrete integration over the edges makes this minimizer invariant of the last shape's parametrization.
Then, the motion trajectory exhibits a \(\chi\)-holonomy of \(g\in G\) if $g=h$.
This is implemented by defining \(h\) through the Euler-Lagrange equation of \eqref{eq:rigidregistration}.

Due to the \(G\)-symmetry of the Lagrangian, motion trajectories are only determined up to a global rigid body transformation. To resolve this ambiguity, we constrain the center of mass of one shape---typically the first in the sequence. However, no constraint on the orientation is necessary: the optimization naturally identifies an optimal initial orientation for achieving the displacement objective.

Moreover, to rule out the trivial solution of a constant path consisting of collapsed bodies for periodic boundary problems, we constrain the edge length of one shape to be strictly positive.
In the experiments we also considered comparisons to kinematic systems, such as Purcell's swimmer, which require \emph{isometry constraints}---additional hard constraints on all or some edge lengths.

\section{Implementation}\label{appsec:Implementation}
We have implemented our method in Python.
To enable the use of second-order optimization methods, we derive explicit expressions for the gradients and Hessians of all functions involved in the optimization. The inner dissipation is implemented in C++ and we use nanobind~\cite{nanobind} to generate Python bindings and SciPy's \texttt{\footnotesize trust-constr} solver~\cite{Virtanen:2020:Scipy} for the constraint optimizations.

\subsection{Initialization and hierarchical refinement}
\label{sec:HierarchicalRefinement}
\begin{figure}[h]
	\centering
	\includegraphics[width = \linewidth]{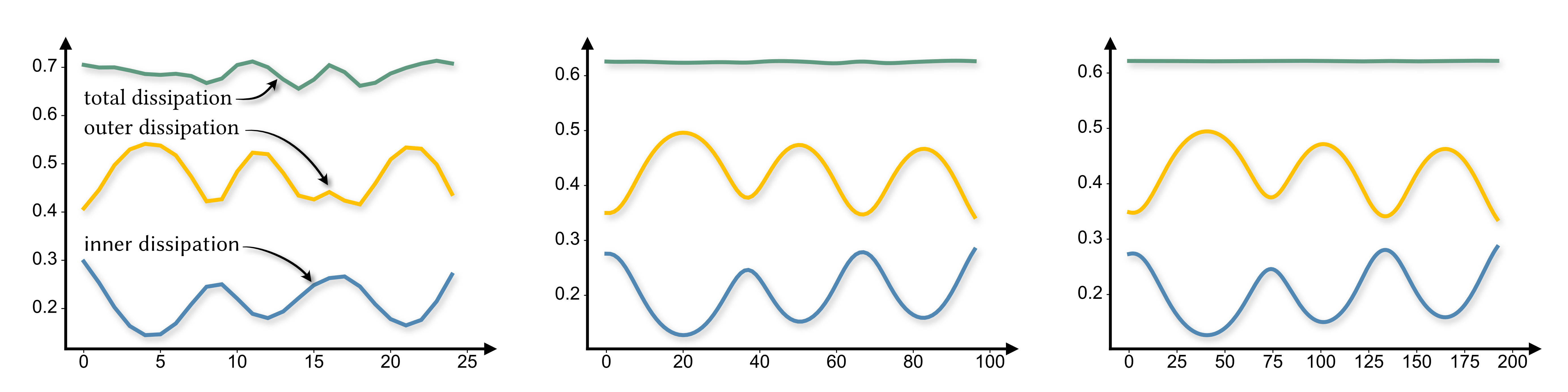}
	\caption{Graphs of the inner, outer and total dissipation of three motion trajectories for increasingly fine time resolutions (left to right: $K=25,\;100,\;200$) of the same geodesic boundary value problem.}
	\label{fig:TimeRefinement}
\end{figure}

To speed up the convergence of our method, we first compute a Riemannian geodesic on the configuration space by solving the unconstrained optimization, since the absence of horizontality constraints significantly reduces computational complexity. 
We then use this path to initialize the constrained optimization.
Moreover, we employ a hierarchical refinement scheme of the time discretization and optimize the motion trajectories across multiple optimization passes, doubling the time-resolution for each subsequent optimization pass. \figref{fig:TimeRefinement} demonstrates that for sufficient resolution, the optimal motion trajectories exhibit the constant total dissipation required of (sub-Riemannian) geodesics.

Although our method identifies solutions that satisfy the optimality conditions (Euler-Lagrange equations) up to machine precision, these solutions are generally only local minima and the solutions might depend on the chosen initial condition for the optimization. 
In our experiments, we observed that especially the number of traversed gait cycles of periodic shape sequences is influenced by the initialization. 
While  changes in the number of traversed gait cycles can be observed
at times, they are rare.

\subsection{Convergence under refinement}
\label{supp-sec:Convergence}
\begin{figure}[h]
	\centering
	\includegraphics[width = .8\linewidth]{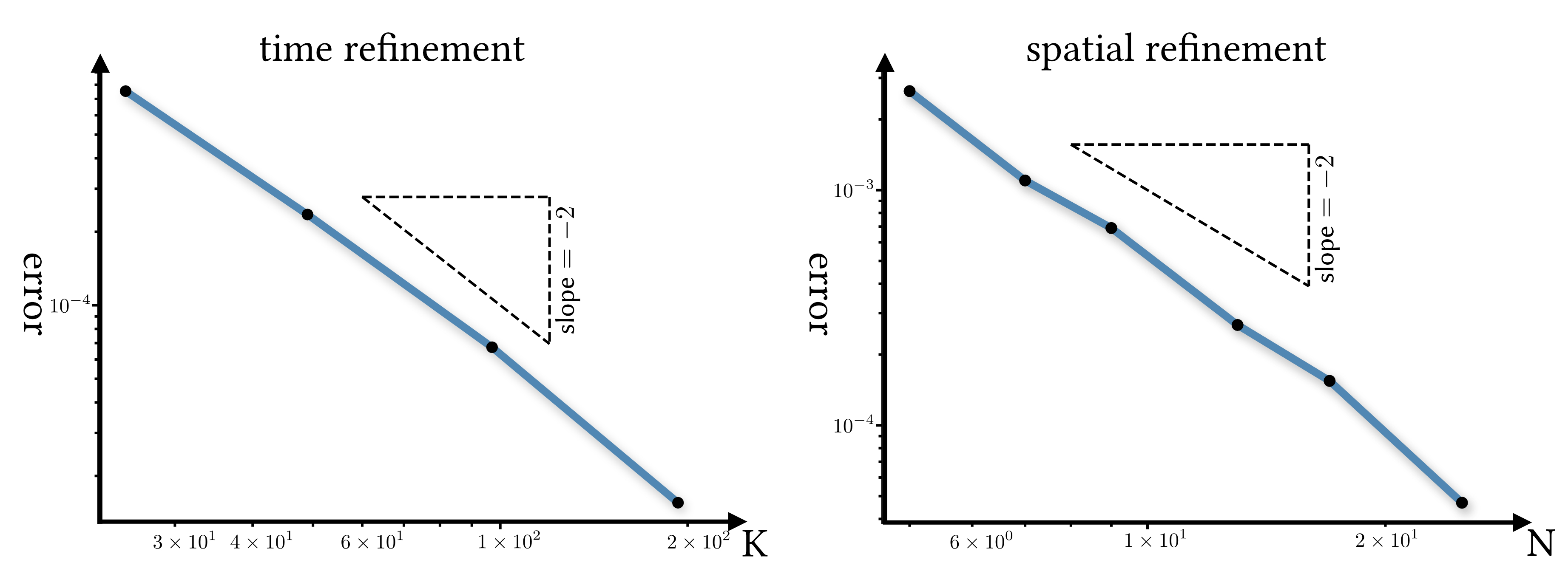}
	\caption{Convergence plots provide numerical evidence of approximately quadratic convergence of the discretization to the continuous limit under time \figloc{(left)} and spatial \figloc{(right)} refinement.}
	\label{fig:TimeAndSpaceConvergence}
\end{figure}

\begin{figure}[h]
	\centering
	\includegraphics[width = .8\linewidth]{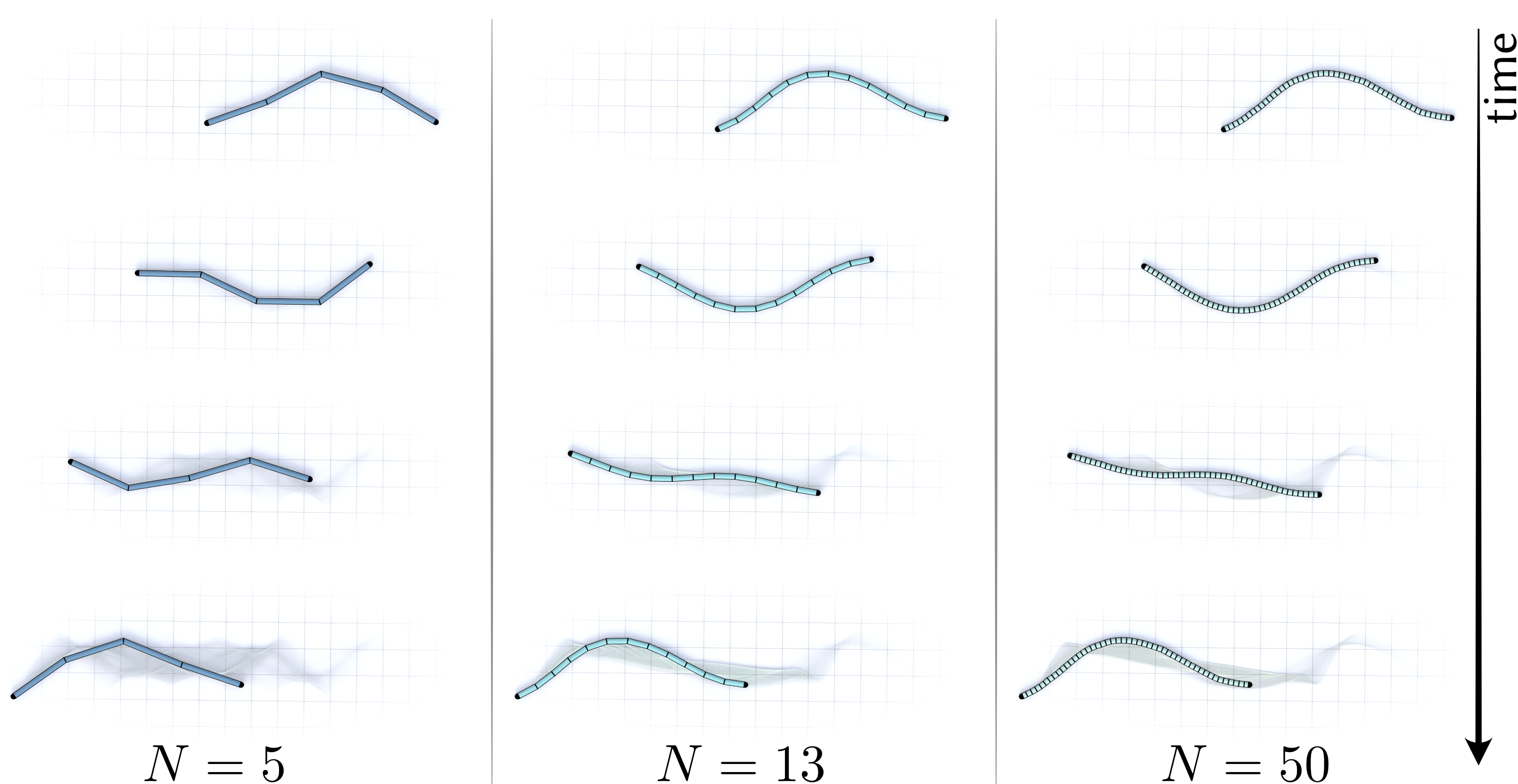}
	\caption{Optimal motion trajectories solving the same geodesic boundary value problem for different spatial resolutions of the locomotor and for a fixed number of time steps.}
	\label{fig:ConvergenceSpatial}
\end{figure}

\begin{figure}[h]
	\centering
	\includegraphics[width = .8\linewidth]{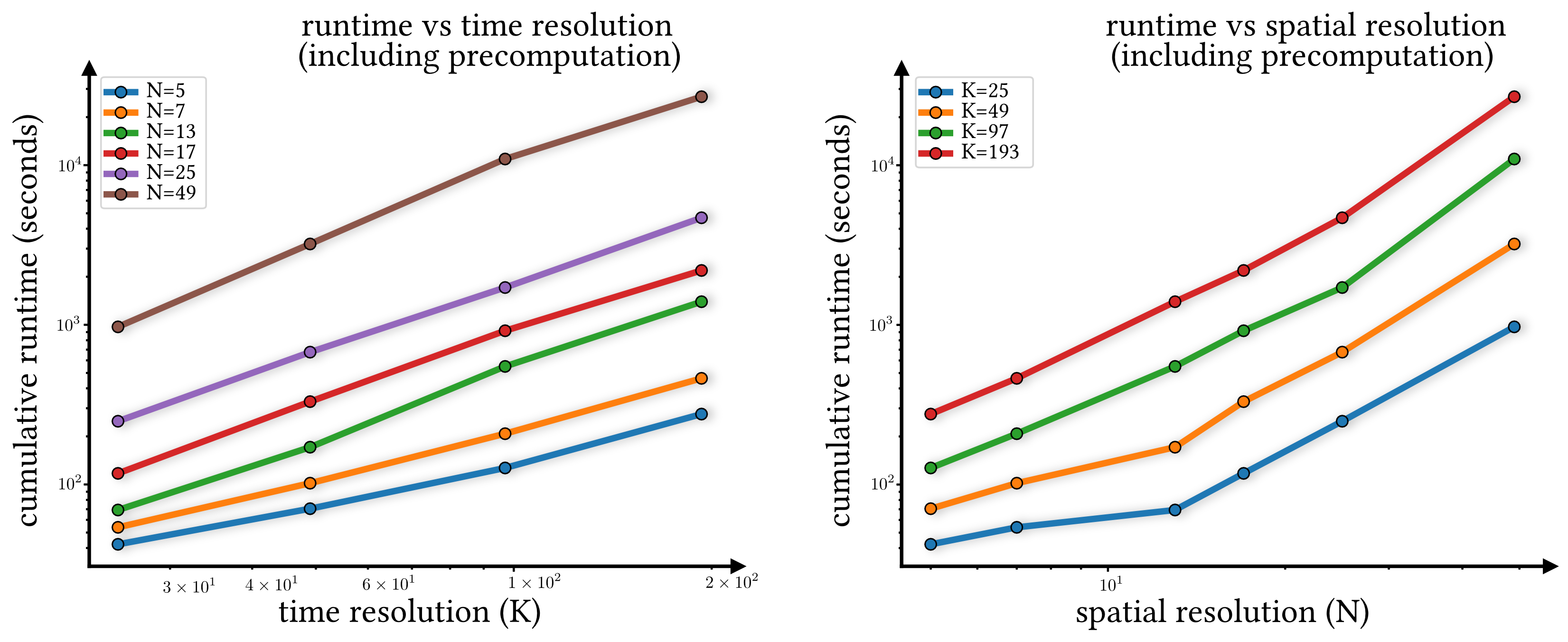}
	\caption{Runtimes for the convergence experiment shown in \figref{fig:TimeAndSpaceConvergence}. \label{fig:runtime}}
\end{figure}

Our experiments show quadratic convergence of the employed discretization scheme to the continuous theory in the limit for both spatial and temporal refinement (\figref{fig:TimeAndSpaceConvergence}). 
\figref{fig:ConvergenceSpatial} shows consistent qualitative behavior of the resulting optimal trajectories under spatial refinement.
We report the corresponding runtimes on a personal computer in \figref{fig:runtime}. 

\section{Comparisons to reduced-order models}
Here we provide further details for the comparison with reduced-order models (Sec.\,3.2. in the main text).
We compare an optimal motion trajectory identified by our model to a gait in the so-called serpenoid shape space. In \cite{Rieser:2024:GPp} the authors found that gaits of natural locomotors at all scales follow approximately circular trajectories in the serpenoid shape space, for an appropriately chosen wavelength. Therefore, motion trajectories integrated from periodic shape sequences that are obtained from uniformly sampling appropriate ellipses in a serpenoid shape space provide a good approximation for ``natural'' gaits.

For a comparison, we first use the method proposed by \cite{Becker:2025:IGL} to identify an optimal ``elliptical'' gait in the serpenoid shape space. Specifically, we optimize parameters for the ellipse's shape, position, orientation as well as the space's wavelength. The objective for the optimization is defined to enforce maximal displacement, regularized with terms accounting for inner and outer dissipation. Next, we solve the corresponding geodesic boundary value problem, using boundary data from the serpenoid gate and initialization with the same gait. This leaves us with two gaits, with the same start and end configurations.  

\begin{figure}[h]
	\centering
	\includegraphics[width = \linewidth]{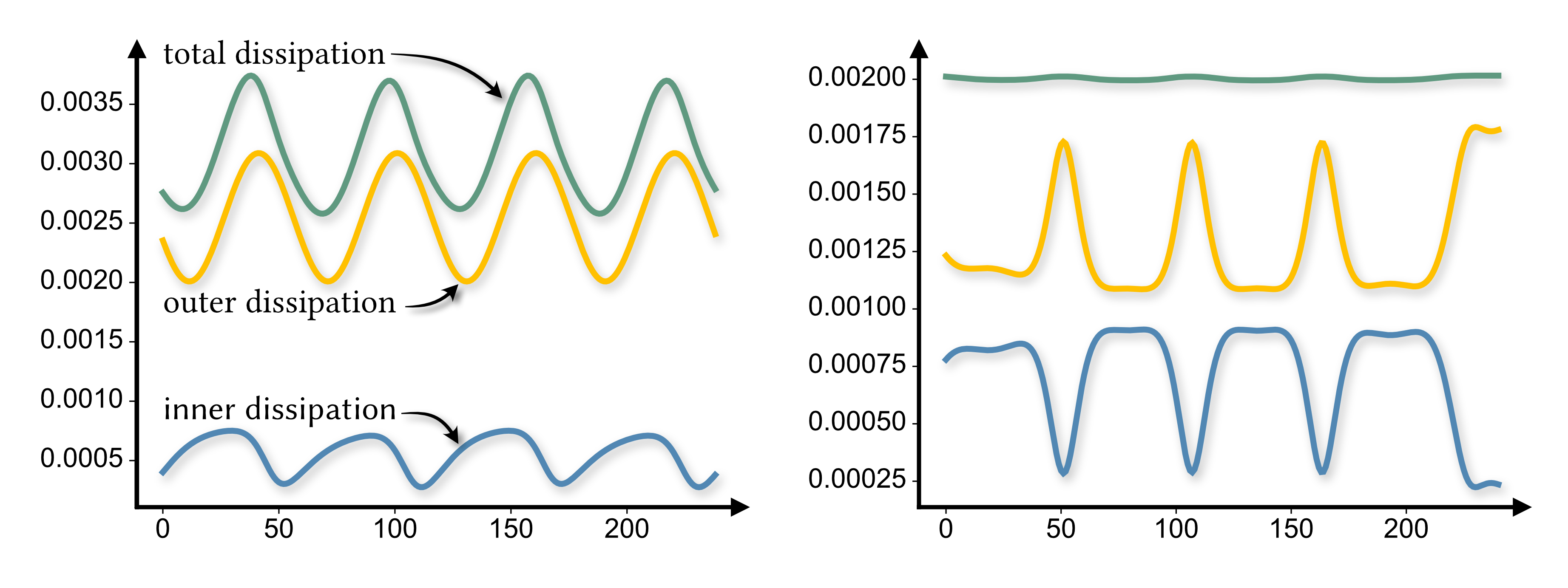}
	\caption{Graphs of the inner, outer and total dissipation of the elliptical gait in the serpenoid shape space \figloc{(left)} and the optimal motion trajectory identified by our method by solving the corresponding geodesic boundary value problem \figloc{(right)}.}
	\label{fig:SerpenoidDissipation}
\end{figure}

Our optimization in the high-dimensional shape space reduces the total energy dissipation (Fig.\,7. in the main text), and leads to a constant total dissipation along the geodesic paths (\figref{fig:SerpenoidDissipation}).

\section{Purcell's swimmer and its generalizations}
\label{supp-sec:Purcell}
\begin{figure}
	\centering
	\includegraphics[width = 0.73\linewidth]{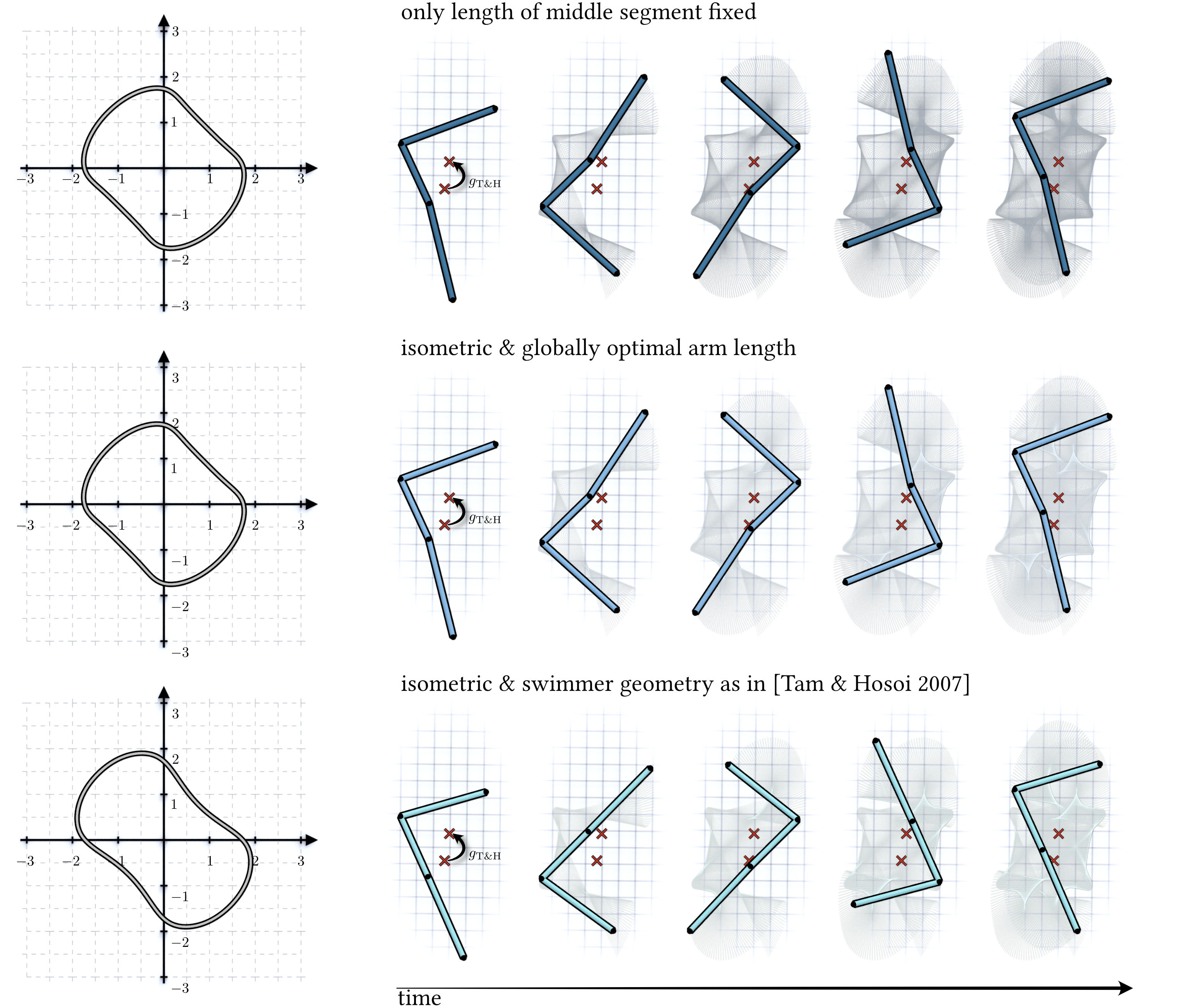}
	\caption{Optimal motion trajectories for different types of Purcell's swimmers visualized in angle coordinates (left) and in configuration space (right). The red crosses indicate the center of mass of initial and target configuration respectively. \figloc{Top:} Stretching/compression of the swimmer's arms over time is allowed, while the length of the middle segment is held fixed. \figloc{Middle:} Isometric deformations for a globally optimized fixed arm length. \figloc{Bottom:} Using the swimmer geometry from  \cite{Tam:2007:OSP} we recover the same optimal gait.
	}
	\label{fig:Purcell_1_0_Displacement}
\end{figure}

This section extends the results in Sec.\,3.3. of the main text by presenting additional experiments on Purcell's swimmer and its generalizations, made possible by our more flexible geometric model assumptions, \ie, allowing for strain distortions. 
In \figref{fig:Purcell_1_0_Displacement} we reproduce the maximal displacement gait \(g_{\text{T\&H}}\) identified by \cite{Tam:2007:OSP} in the absence of inner dissipation for constant edge lengths and compare it to scenarios with fixed optimal, and time-dependent edge lengths.

\begin{figure}
	\centering
	\includegraphics[width = 0.73\linewidth]{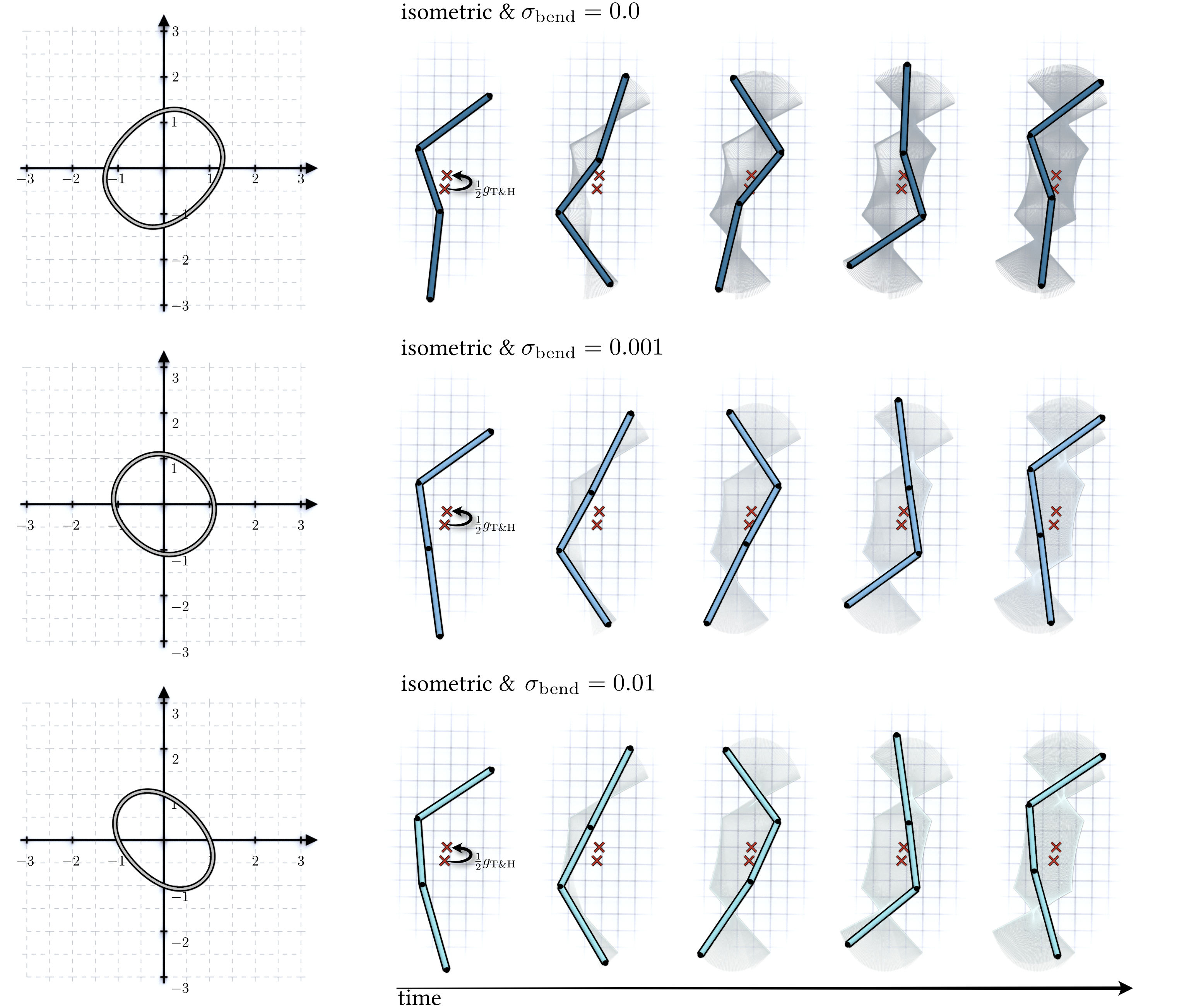}
	\caption{Influence of the bending energy on optimal motion trajectories of Purcell's swimmer (isometric) for different weights \(\sigma_{\rm bend}\) with a fixed displacement equal to the half of the maximal displacement \(g_{\text{T\&H}}\) identified by \cite{Tam:2007:OSP}.}
	\label{fig:Purcell_0_5_Displacement}
\end{figure}

\figref{fig:Purcell_0_5_Displacement} shows an ablation study on the influence of the bending energy on its optimal motion trajectories. Here, we fix the swimmers' edge lengths and solve periodic boundary value problems for different weights \(\sigma_{\rm bend}\). The target displacement is fixed to half the maximal displacement \(g_{\text{T\&H}}\) identified by \cite{Tam:2007:OSP}.

\begin{figure}
	\centering
	\includegraphics[width = 0.85\linewidth]{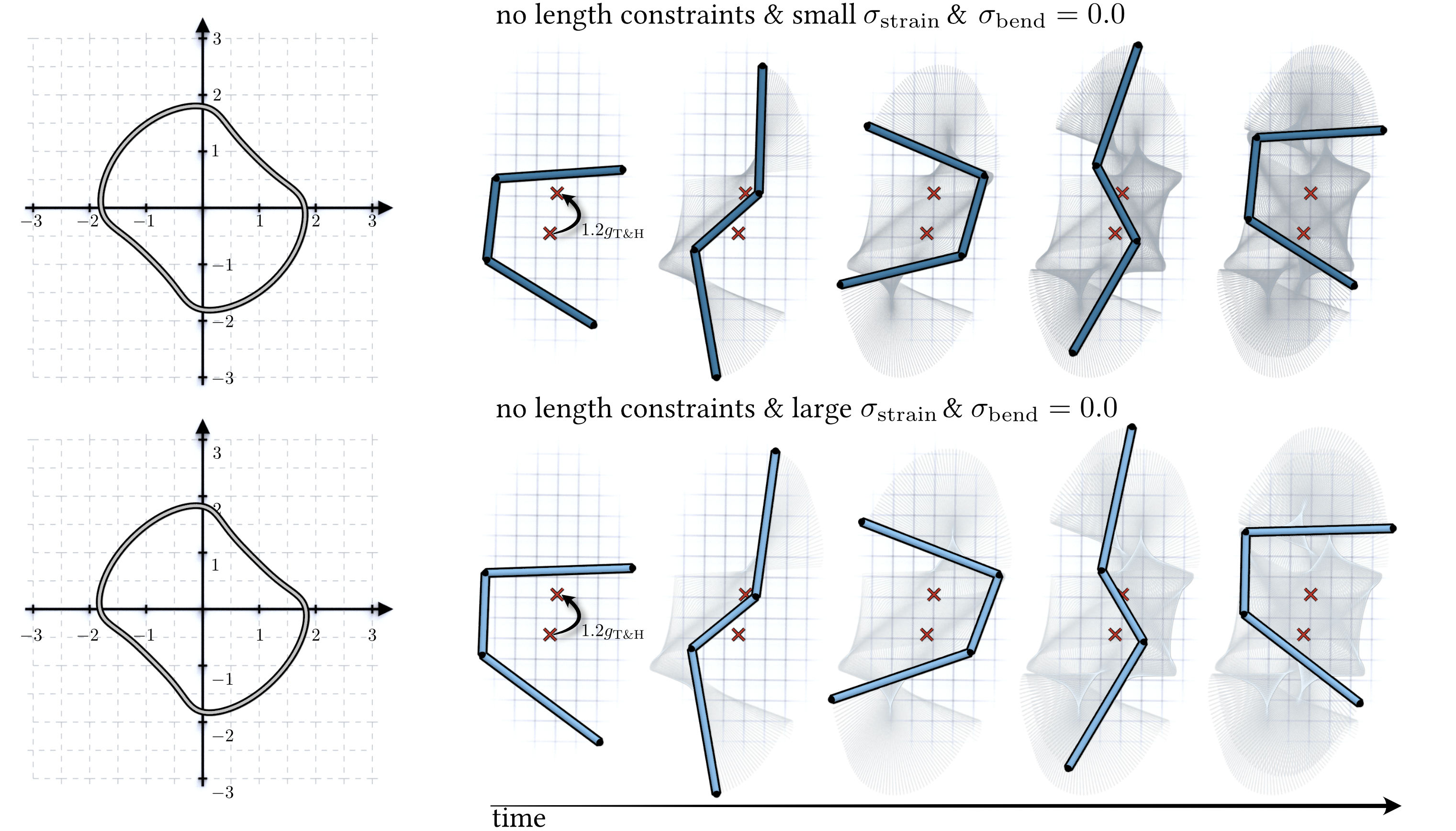}
	\includegraphics[width = 0.85\linewidth]{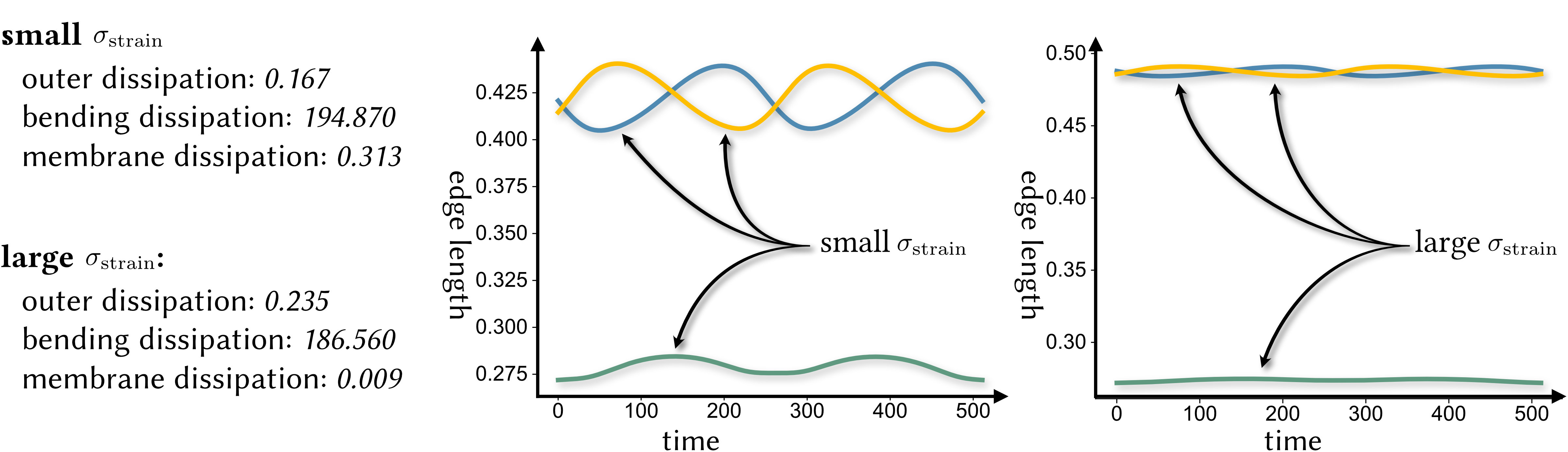}
	\caption{Optimal motion trajectories for a generalized
		Purcell's swimmer, achieving a displacement of \(1.2\) times
		the maximal displacement \(g_{\text{T\& H}}\) identified by
		\cite{Tam:2007:OSP}, obtained without applying penalties on
		the bending component of the inner dissipation
		(\(\sigma_{\rm bend}=0\)). \figloc{Top row:} Small strain weight \(\sigma_{\rm strain}\). \figloc{Middle row:} Large strain weight \(\sigma_{\rm strain}\). \figloc{Bottom row:} Numerical values for inner and outer dissipation \figloc{(left)} and plots of the swimmers' edge lengths (green middle edge, blue and yellow outer edges) over time for small \figloc{(middle)} and large \figloc{(right)} strain weights \(\sigma_{\rm strain}\).
		\label{fig:Purcell_1_2_Displacement}}
\end{figure}

Complementing the above experiment, \figref{fig:Purcell_1_2_Displacement} explores the influence of the strain energy in the absence of bending penalty. Notably, small weights on the strain energy lead to optimal motion trajectories that exhibit significant periodic distortion of the swimmers' edge lengths. In contrast, large weights lead to merely small deformations of the edge lengths. For both cases the motion trajectory remains qualitatively similar.

\begin{figure}
	\centering
	\includegraphics[width = 0.85\linewidth]{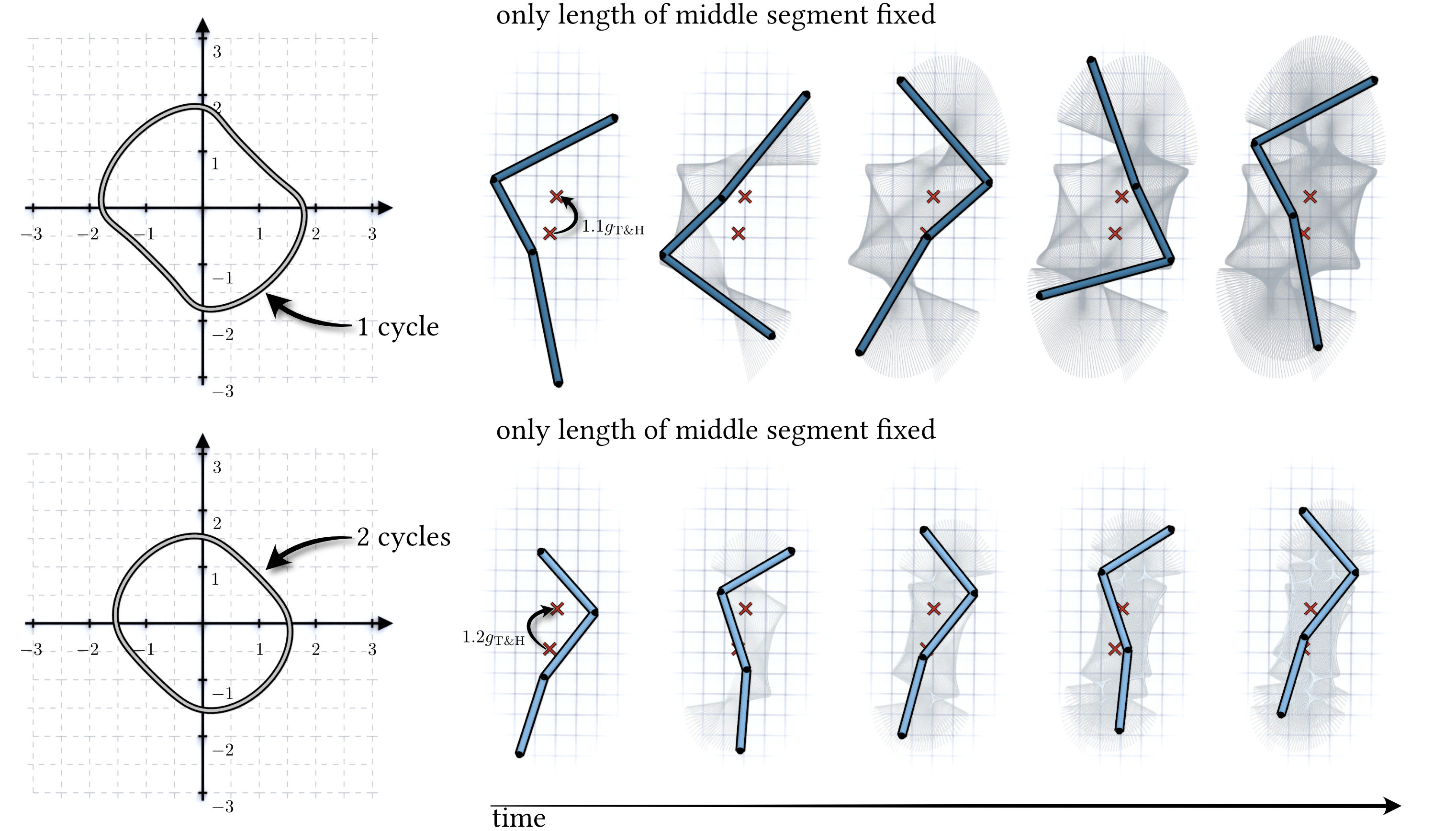}
	\caption{Optimal motion trajectories for a generalized Purcell's swimmer with fixed length of middle segment, displacing further than the maximal displacement \(g_{\text{T\&H}}\) identified by \cite{Tam:2007:OSP}, while allowing for stretching/compression of the outer edges, and no penalty on bending. \figloc{Top:} \(1.1\, g_{\text{T\&H}}\) displacement traversing one gait cycle. \figloc{Bottom:} \(1.2\, g_{\text{T\&H}}\) displacement traversing two gait cycles.}
	\label{fig:Purcell_1_1_Displacement}
\end{figure}

Finally, \figref{fig:Purcell_1_1_Displacement} demonstrates that allowing time-dependent distortions of the swimmer's edge lengths enhances its performance. Our generalized Purcell's swimmer outperforms the maximal displacement gait identified by \cite{Tam:2007:OSP}, achieving a displacement of \(1.1\,g_{\text{}T\& H}\) with a single gait cycle. Note that this is not a precise upper bound for the displacement achievable in one gait. However, our experiments indicate that for a displacement of \(1.2\,g_{\text{}T\& H}\) our optimization converges to a solution that requires traversing a cycle twice.

\end{document}